\documentclass[lettersize,journal]{IEEEtran}
\usepackage{amsmath,amsfonts}
\usepackage{algorithmic}
\usepackage{algorithm}
\usepackage{array}
\usepackage[caption=false,font=normalsize,labelfont=sf,textfont=sf]{subfig}
\usepackage{textcomp}
\usepackage{stfloats}
\usepackage{url}
\usepackage{verbatim}
\usepackage{graphicx}
\usepackage{cite}
\usepackage{tabularx}
\usepackage{booktabs} % 确保 \toprule, \midrule, \bottomrule 正常工作
\usepackage{xcolor}
\usepackage{booktabs}
\usepackage{multirow}
\usepackage{graphicx}
\usepackage[table]{xcolor} % 必须加上 [table]，用于支持 \cellcolor
\usepackage{booktabs}
\usepackage{multirow}
\usepackage{graphicx}

\begin{document}

\title{AquaFlow: A Monocular Gaussian Splatting SLAM for \\ Underwater Streaming Reconstruction}
% AquaFlow: Underwater Monocular Gaussian Splatting SLAM  \\ via Hybrid Scene Modeling


\author{
Yingxiang Xu$^*$, 
Kerui Ren$^*$,
Wenqi Guo,
Changjian Jiang,
Tao Lu,
Linning Xu,
Mulin Yu$^\dag$
% <-this % stops a space
\thanks{Y. Xu is with Zhejiang University and Shanghai AI Laboratory. E-mail: xuyingxiang@pjlab.org.cn.}
\thanks{K. Ren is with Shanghai Jiao Tong University and Shanghai AI Laboratory. E-mail: renkerui@sjtu.edu.cn.}
\thanks{W. Guo is with Tsinghua University and Shanghai AI Laboratory. E-mail: guowenqi@pjlab.org.cn.}
\thanks{C. Jiang is with The University of Hong Kong. E-mail: changjianjiang01@gmail.com.}
\thanks{T. Lu is with Shanghai AI Laboratory. E-mail: lutao@pjlab.org.cn}
\thanks{L. Xu is with The Chinese University of Hong Kong. E-mail: linningxu@link.cuhk.edu.hk.}
\thanks{M. Yu is with Shanghai AI Laboratory. E-mails: yumulin@pjlab.org.cn.}
\thanks{$*$ Equal contribution.}
\thanks{$\dag$ Corresponding author.}
}

% \markboth{IEEE TRANSACTIONS ON MULTIMEDIA}{IEEE TRANSACTIONS ON MULTIMEDIA}%

\maketitle

\begin{abstract}
Recent monocular 3D Gaussian Splatting (3DGS) streaming reconstruction methods have achieved impressive performance by balancing reconstruction quality and efficiency. 
% However, extending these frameworks to underwater scenes remains challenging due to severe visual degradation that affects camera pose tracking and water-induced image formation effects, such as attenuation and scattering, that degrade scene reconstruction.
However, extending these frameworks to underwater scenes remains challenging due to severe visual degradation, such as light attenuation and scattering, which degrades camera pose tracking and distorts scene geometry.
To address these challenges, we propose AquaFlow, a monocular Gaussian Splatting streaming reconstruction framework for efficient and high-fidelity underwater reconstruction. Specifically, AquaFlow fine-tunes a 3D vision foundation model on large-scale underwater data for robust pose and pointmap estimation, and introduces a medium-guided incremental Gaussian initialization strategy for streaming mapping. Furthermore, we develop a streaming-compatible hybrid scene representation that integrates structured, distance-conditioned neural Gaussians with a physics-inspired optical model to compensate for underwater image formation effects, enabling accurate scene reconstruction. We evaluate AquaFlow on a comprehensive dataset of 62 diverse underwater trajectories, collected from both public benchmarks and in-the-wild web videos across various scales. Extensive experiments demonstrate that AquaFlow achieves state-of-the-art tracking and rendering performance, reducing average localization error by 13.2\% and improving PSNR by 4.74 dB compared to WaterSplat-SLAM.
\end{abstract}


\section{Introduction}

\IEEEPARstart{H}{igh-resolution}, photorealistic underwater 3D reconstruction is fundamental to benthic habitat mapping, underwater resource assessment, marine infrastructure inspection, and autonomous subsea navigation~\cite{song2022optical,wolfl2019seafloor,wang2023real,bodenmann2017methods,hernandez2016autonomous}. 
Yet, scalable and efficient 3D reconstruction from continuous underwater video remains severely underexplored. 
This slow progress is primarily because complex optical attenuation and scattering induce severe visual degradation, making it extremely challenging to rapidly estimate stable camera poses and achieve high-accuracy scene reconstruction within short timeframes~\cite{hu2023overview}. 
Despite these challenges, monocular cameras remain a highly desirable solution: compared to bulky sonar or multi-sensor suites, they serve as an exceptionally lightweight, cost-effective, and flexible sensing modality that captures rich visual textures while minimizing vehicle weight, drag, and hydrodynamic impact~\cite{qin2022survey,ferrera2019real,chemisky2021portable,joshi2022high}.

3D Gaussian Splatting (3DGS)~\cite{kerbl20233d} provide an efficient representation for high-fidelity rendering and open new opportunities for underwater reconstruction. 
However, existing underwater 3DGS methods are predominantly offline~\cite{li2024watersplatting,yang2024seasplat,jiang2025rusplatting,xing2025uw3dgs}, requiring complete image sequences alongside pre-estimated camera poses and sparse point clouds as input, while incurring severe computational overheads for Gaussian optimization and densification even setting aside the lengthy preprocessing time of Structure-from-Motion (SfM).
Moreover, existing methods are predominantly evaluated on a limited number of small-scale scenes, leaving their scalability to extended video sequences and their generalizability across diverse, complex underwater topologies largely unproven. 
% Streaming reconstruction addresses this limitation by incrementally estimating camera motion and updating the scene representation as new frames arrive, making it well suited to continuous video feeds and long-duration subsea missions.

To overcome these efficiency and scalability bottlenecks, streaming reconstruction has emerged as a promising paradigm, with recent advances in 3DGS-based SLAM~\cite{meuleman2025fly,zhang2025hi,cheng2025outdoor} providing a solid technical foundation to incrementally update explicit scene representations for time-efficient mapping and high-quality novel-view synthesis. However, these methods are primarily tailored for terrestrial environments, and directly applying them to underwater scenarios introduces two tightly coupled challenges: robust pose tracking and medium-aware scene modeling. First, wavelength-dependent attenuation, non-uniform illumination, and weakly textured sandy or rocky seafloors substantially reduce cross-frame visual correspondence. The resulting pose errors and trajectory drift destabilize dense point-map initialization and propagate into the Gaussian representation, ultimately degrading both geometry and rendering quality. Although recent SLAM systems~\cite{murai2025mast3rslam,maggio2026vggt,maggio2026vggt2} employ 3D vision foundation models to improve visual matching, their predominantly terrestrial training data limit their ability to generalize to subsea imagery governed by complex light transport. Second, the clear-medium assumption underlying standard 3DGS breaks down underwater, where image formation is strongly affected by distance-dependent spectral attenuation and volumetric backscattering~\cite{levy2023seathru}. Without modeling these effects, 3DGS primitives absorb them into static Gaussian colors, leading to geometric distortions and rendering artifacts.

\begin{figure*}[t]
  \centering
  \includegraphics[width=\textwidth]{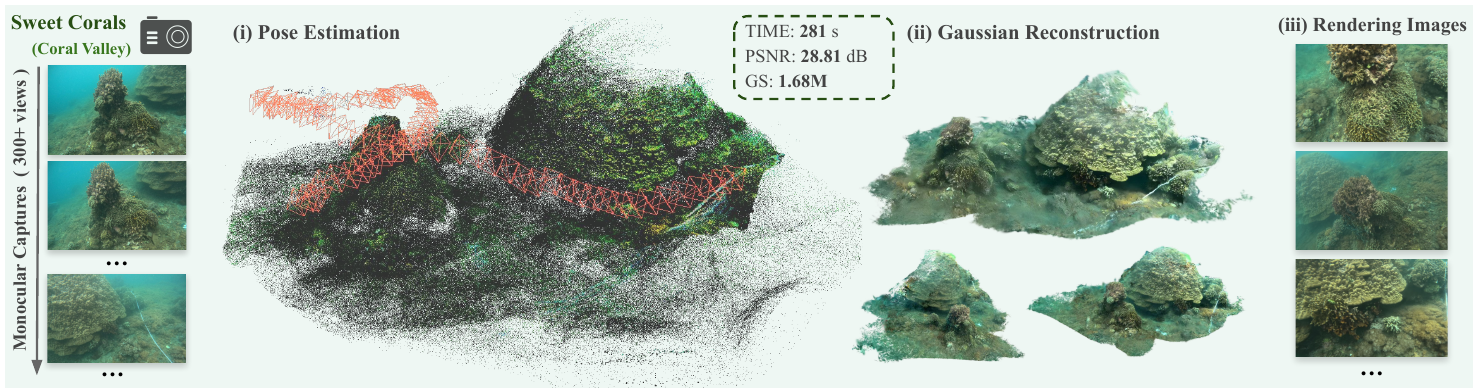}
  \caption{\textbf{Overview of the AquaFlow framework.} Given a long, continuous underwater video sequence, AquaFlow performs online streaming 3D reconstruction, rapidly estimating camera poses, dense pointmaps, a highly compact 3D Gaussian map, and physical medium parameters. To achieve this, AquaFlow seamlessly integrates domain-adapted underwater 3D geometric priors, a medium-guided Gaussian initialization scheme, and a hybrid scene representation combining structured neural Gaussians with physics-based medium modeling, ultimately enabling accurate subsea structural recovery and photorealistic novel view synthesis.}
  \label{fig:teaser}
\end{figure*}

To address these challenges, we present \textbf{AquaFlow}, a monocular 3D Gaussian Splatting SLAM framework that combines underwater-adapted geometric priors with physics-guided hybrid scene modeling for scalable, high-fidelity subsea reconstruction (Fig.~\ref{fig:teaser}). For robust tracking, AquaFlow fine-tunes a 3D vision foundation model on large-scale subsea corpora, thereby distilling robust geometric priors for accurate pose tracking and dense point-map estimation. For online mapping, we develop a streaming-compatible hybrid scene-medium representation that integrates distance-aware neural Gaussian primitives with an explicit underwater image formation model. During streaming, the physical model guides medium-aware Gaussian initialization in unobserved regions, while distance-encoded Gaussian parameters adaptively capture fine-grained underwater imaging effects not fully represented by the explicit formulation. By unifying explicit optical modeling with implicit neural representations, AquaFlow decouples intrinsic scene geometry from complex optical degradation, enabling highly accurate and photorealistic subsea mapping.

Our main contributions are summarized as follows:
\begin{itemize}
    \item We fine-tune a 3D vision foundation model on extensive subsea datasets to bridge the terrestrial-to-underwater domain gap, providing robust geometric priors that empower camera pose tracking and dense depth initialization.

    \item We design a streaming-compatible hybrid scene representation combining distance-aware neural Gaussians with a physical underwater optical model, effectively disentangling intrinsic appearance from distance-dependent attenuation and backscattering.

    \item We compile a comprehensive underwater evaluation benchmark containing 62 diverse real-world sequences (25 from public datasets and 37 challenging in-the-wild subsea videos). Extensive experiments validate that AquaFlow achieves state-of-the-art accuracy in both camera tracking and high-fidelity 3D reconstruction.
\end{itemize}

\section{Relate works}
\subsection{Underwater Visual SLAM and Pose Estimation}
To tackle severe image degradation, scattering, and low illuminance in underwater environments, classical visual SLAM systems often suffer from feature tracking failures or misalignments. Existing studies primarily address these issues from a 2D perspective by applying self-adaptive color or turbidity restoration\cite{chen2025underwater} and lightweight GANs\cite{zheng2023real} as pre-processing backbones to restore image quality, replacing conventional hand-crafted features with domain-tailored learned descriptors\cite{yang2023knowledge,wang2025nerf}, or leveraging deep optical flow and spatial attention mechanisms\cite{wang2025eum,wu2025raem} to adapt to weakly textured underwater regions. However, these 2D methods only yield sparse point clouds and coarse scene geometry, while neglecting the strong generalizability and dense 3D structure prediction capabilities of modern 3D vision foundation models for downstream matching and pose estimation in degraded underwater regions.

\subsection{Offline Underwater 3D Gaussian Splatting Reconstruction}
To achieve photorealistic novel view synthesis, recent efforts have extended 3DGS to underwater environments by jointly optimizing Gaussian primitives and explicit physical light transport models from pre-computed Structure-from-Motion (SfM) inputs. For instance, methods like Water-Splatting~\cite{li2024watersplatting} and SeaSplat~\cite{yang2024seasplat} integrate image formation models into 3DGS to disentangle scene radiance from attenuation and backscatter, while UW-GS~\cite{wang2025uw} and RUSplatting~\cite{jiang2025rusplatting} incorporate depth- or range-dependent priors to restore appearance and geometry. However, as scene scale expands, these offline paradigms demand excessive computation time for offline Gaussian densification, pruning, and optimization, failing to accommodate the continuous streaming video nature of practical underwater data collection. Furthermore, they heavily rely on simplified physical medium formulations that fail to capture complex underwater imaging dynamics, and are typically evaluated only on narrow, small-scale datasets with sparse frame counts.

\begin{figure*}[t]
  \centering
  \includegraphics[width=\textwidth]{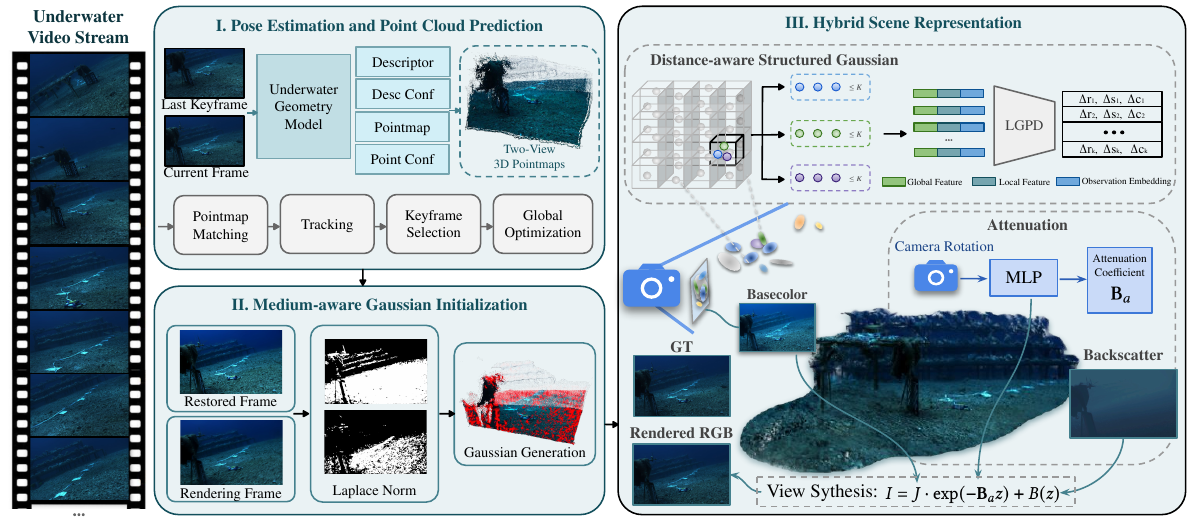}
  \caption{\textbf{Overview of the AquaFlow streaming 3D reconstruction pipeline.} 
  Given incoming monocular underwater frames, AquaFlow first predicts dense pointmaps and descriptors using a \textbf{domain-adapted 3D foundation model} to establish reliable pixel correspondences for relative pose optimization. 
  Based on match counts and parallax, frames are dynamically categorized into keyframes, mapping frames, or common frames. 
  Keyframes and mapping frames trigger \textbf{medium-guided incremental initialization}, where an online learned physical medium model removes attenuation and backscatter to yield clean base colors and structure-aware Gaussian seeds. 
  Finally, these primitives are organized into a \textbf{distance-aware structured neural Gaussian representation} and decoded via the LGPD module, which is jointly optimized alongside explicit physical medium modeling for high-fidelity rendering.}
  \label{fig:pipeline}
\end{figure*}

\subsection{Streaming Gaussian Splatting Reconstruction}
Combining 3DGS with SLAM enables real-time state estimation to guide Gaussian initialization and rapid scene learning, presenting a promising paradigm for online underwater reconstruction\cite{matsuki2024gaussian,deng2024compact,sandstrom2025splat,li2026artdeco}. However, current general streaming reconstruction algorithms still face significant challenges when applied to underwater environments, including severe pose drift, erroneous Gaussian growth, and suboptimal scene representation. Meanwhile, WaterSplat-SLAM\cite{wang2026watersplat}, the concurrent underwater streaming method, attempts to address underwater degradation by leveraging pre-computed semantic segmentation to guide frame tracking and Gaussian optimization. Nevertheless, its heavy reliance on semantic masks renders it vulnerable to segmentation failures and computationally unfeasible in complex or low-visibility underwater scenes.

% To overcome these limitations, we propose AquaFlow, a novel framework that pioneers the integration of generalizable underwater 3D priors into a streaming-tailored hybrid scene representation. Extensive evaluations across diverse sequence topologies validate the exceptional generalization and robust reconstruction capabilities of our priors and hybrid representation.

\section{Methodology}
\label{sec:method}

We address the problem of incrementally recovering a high-fidelity 3D underwater scene from a single monocular RGB video. Given an incoming sequence $\{I_t\}_{t=1}^{N}$, our objective is to jointly estimate the camera trajectories $\{\mathbf{T}_t\}_{t=1}^{N} \in SE(3)$, construct a dense Gaussian scene map $\{\mathcal{G}_j\}_{j=1}^{M}$, and resolve the optical parameters governing underwater medium attenuation and backscatter. 

To overcome severe domain shifts and complex medium-induced optical degradation inherent in underwater environments, AquaFlow decomposes the streaming reconstruction process into four synergistic modules: 1) an underwater-adapted 3D geometric prior (Sec.~\ref{sec:prior}), 2) tracking and global optimization (Sec.~\ref{sec:pose}), 3) a hybrid scene-medium neural representation (Sec.~\ref{sec:representation}), and 4) an explicit joint optimization objective (Sec.~\ref{loss_function}).

\subsection{Generalizable Underwater 3D Prior}
\label{sec:prior}

\subsubsection{Data Collection and Processing}
To address the lack of generalizable multi-view geometric foundation models for underwater environments, we construct the first large-scale, highly diverse underwater 3D dataset tailored for model adaptation. Spanning a wide variety of synthetic and real-world underwater environments, including ocean floors, coral reefs, canyons, and submerged structures—our dataset provides rich geometric supervision.

For synthetic data~\cite{wang2020tartanair,alvarez2023mimir}, camera poses and depth maps are directly extracted from simulator annotations. For real-world sequences, we employ domain-specific strategies: \textit{sweet-corals}~\cite{wildflow_sweet_corals_indo_2025} is reconstructed via COLMAP-based photogrammetry~\cite{schoenberger2016sfm}, while \textit{FLSea-stereo}~\cite{randall2023flsea} combines stereo depth estimation with feature-based pose-only bundle adjustment. All image-depth pairs are undistorted and calibrated to enforce rigid geometric consistency. To eliminate noisy supervision, we apply joint geometric and semantic masking to remove dynamic objects, invalid depth regions, specular highlights, and scattering artifacts. Finally, valid pixels are back-projected to evaluate cross-view spatial overlap; only frame pairs with sufficient overlap ($\ge 30\%$) and verified geometric consistency are retained for training.

\subsubsection{Model finetuning}
% To establish a robust 3D vision foundation model tailored for underwater environments, we select MASt3R\cite{leroy2024grounding} as our base architecture due to its rapid inference speed, exceptional stability in small-baseline geometric estimation, and strong alignment with front-end tracking requirements in SLAM frameworks. Given a pair of uncalibrated underwater images $I_i, I_j \in \mathbb{R}^{H \times W \times 3}$, MASt3R predicts the 3D geometry, per-pixel descriptors, and their associated confidence maps in a single forward pass:
% \begin{equation}
% \label{eq:mast3r_output}
% (X_i^i, X_i^j, D_i^i, D_i^j, Q_i^i, Q_i^j, C_i^i, C_i^j) = f_{\theta}(I_i, I_j)
% \end{equation}
% where $X_i^i, X_i^j \in \mathbb{R}^{H \times W \times 3}$ denote the dense 3D pointmaps of images $I_i$ and $I_j$ in the coordinate frame of camera $i$. The terms $D_i^i, D_i^j \in \mathbb{R}^{H \times W \times d}$ represent the per-pixel descriptors for images $I_i$ and $I_j$, respectively. To handle underwater noise, $Q_i^i, Q_i^j \in \mathbb{R}^{H \times W \times 1}$ provide confidence estimates for the predicted 3D point coordinates, while $C_i^i, C_i^j \in \mathbb{R}^{H \times W \times 1}$ represent the descriptor confidences for feature matching.

To establish a robust 3D vision foundation model tailored for underwater environments, we select MASt3R~\cite{leroy2024grounding} as our base architecture due to its rapid inference speed, exceptional stability in small-baseline geometric estimation, and strong alignment with front-end tracking requirements in SLAM frameworks. Given a pair of uncalibrated underwater images $\mathbf{I}_i, \mathbf{I}_j \in \mathbb{R}^{H \times W \times 3}$, MASt3R predicts the 3D geometry, per-pixel descriptors, and their associated confidence maps in a single forward pass:
\begin{equation}
\label{eq:mast3r_output}
(\mathbf{X}_i^i, \mathbf{X}_i^j, \mathbf{D}_i^i, \mathbf{D}_i^j, \mathbf{Q}_i^i, \mathbf{Q}_i^j, \mathbf{C}_i^i, \mathbf{C}_i^j) = f_{\boldsymbol{\theta}}(\mathbf{I}_i, \mathbf{I}_j),
\end{equation}
where $\mathbf{X}_i^i, \mathbf{X}_i^j \in \mathbb{R}^{H \times W \times 3}$ denote the dense 3D pointmaps of images $\mathbf{I}_i$ and $\mathbf{I}_j$ expressed in the local coordinate frame of camera $i$. The terms $\mathbf{D}_i^i, \mathbf{D}_i^j \in \mathbb{R}^{H \times W \times d}$ represent the per-pixel feature descriptor maps for images $\mathbf{I}_i$ and $\mathbf{I}_j$, respectively. To handle underwater noise, $\mathbf{Q}_i^i, \mathbf{Q}_i^j \in \mathbb{R}^{H \times W \times 1}$ provide local confidence estimates for the predicted 3D point coordinates, while $\mathbf{C}_i^i, \mathbf{C}_i^j \in \mathbb{R}^{H \times W \times 1}$ represent the descriptor confidences used for robust feature matching.

We fine-tune MASt3R on our dataset of \textbf{224K calibrated underwater two-view image pairs}. We omit scale normalization to directly optimize metric-scale 3D predictions and the overall objective combines 3D regression and feature matching:
\begin{equation}
\mathcal{L}_{\mathrm{total}} = \mathcal{L}_{\mathrm{conf}} + \beta \mathcal{L}_{\mathrm{match}}
\end{equation}
where $\mathcal{L}_{\text{conf}}$ optimizes metric pointmap regression along with a log-confidence regularization term. The matching objective $\mathcal{L}_{\text{match}}$ leverages an InfoNCE contrastive loss over ground-truth point correspondences established according to their proximity in 3D space\cite{leroy2024grounding}.

% =========================================================
% 3.2 Pose Estimation and Point Cloud Prediction
% =========================================================
\subsection{Tracking and GLobal Optimization}
\label{sec:pose}

% For an incoming frame $I_t$, we first execute $f_{\theta}(I_t, I_k)$ by passing $I_t$ together with the latest keyframe $I_k$ into our fine-tuned foundation model. This feed-forward inference directly yields the dense pointmaps $(X_t^t, X_t^k)$, per-pixel descriptors $(D_t^t, D_t^k)$, and their associated confidence maps in the coordinate frame of camera $t$. To establish a set of reliable pixel correspondences $\mathcal{M}_{t,k}$ between $I_t$ and $I_k$, we perform dense descriptor matching using $D_t^t, D_t^k$ combined with reciprocal ray-error optimization over predicted 3D geometries \cite{murai2025mast3rslam}. Subsequently, the relative transformation $\mathbf{T}_{kt} \in \mathrm{Sim}(3)$ is estimated by minimizing the confidence-weighted pointmap alignment residual:
% \begin{equation}
% \mathbf{T}_{kt}^{*} = \arg\min_{\mathbf{T}_{kt}} \sum_{(u,v)\in \mathcal{M}_{t,k}} \rho \left( \left\| X_k^k(v) - \mathbf{T}_{kt} X_t^t(u) \right\|_2 \right),
% \label{eq:pose_opt}
% \end{equation}
% where $X_t^t(u)$ and $X_k^k(v)$ denote the predicted 3D point coordinates at matched pixel positions $u$ and $v$, respectively and $\rho(\cdot)$ represents the Huber robust loss function.

With the underwater 3D prior, the tracking frontend estimates relative transformations and predicts dense point clouds as new frames arrive in a stream. For an incoming frame $\mathbf{I}_t$, we first execute $f_{\boldsymbol{\theta}}(\mathbf{I}_t, \mathbf{I}_k)$ by passing $\mathbf{I}_t$ together with the latest keyframe $\mathbf{I}_k$ into our fine-tuned foundation model. This feed-forward inference directly yields the dense pointmaps $(\mathbf{X}_t^t, \mathbf{X}_t^k)$, per-pixel descriptors $(\mathbf{D}_t^t, \mathbf{D}_t^k)$, and their associated confidence maps in the coordinate frame of camera $t$. To establish a set of reliable pixel correspondences $\mathcal{M}_{t,k}$ between $\mathbf{I}_t$ and $\mathbf{I}_k$, we perform dense descriptor matching using $\mathbf{D}_t^t, \mathbf{D}_t^k$ combined with reciprocal ray-error optimization over predicted 3D geometries \cite{murai2025mast3rslam}. Subsequently, the relative transformation $\mathbf{T}_{kt} \in \mathrm{Sim}(3)$ is estimated by minimizing the confidence-weighted pointmap alignment residual:
\begin{equation}
\mathbf{T}_{kt}^{*} = \arg\min_{\mathbf{T}_{kt}} \sum_{(u,v)\in \mathcal{M}_{t,k}} \rho \left( \left\| \mathbf{X}_k^k(v) - \mathbf{T}_{kt} \mathbf{X}_t^t(u) \right\|_2 \right),
\label{eq:pose_opt}
\end{equation}
where $\mathbf{X}_t^t(u)$ and $\mathbf{X}_k^k(v)$ denote the predicted 3D point coordinates at matched pixel positions $u$ and $v$, respectively and $\rho(\cdot)$ represents the Huber robust loss function.

% Specifically, \textbf{keyframes} are spawned when the size of the reliable match set $|\mathcal{M}_{t,k}|$ falls below a predefined tracking threshold $\tau_{\text{keyframe}}$. This triggers the insertion of $I_t$ into the backend pose graph to jointly optimize all absolute keyframe poses $\{\mathbf{T}_i\}_{i \in \mathcal{K}}$ across active spatial edges $(i,j) \in \mathcal{E}$ via:
% \begin{equation}
%     \{\mathbf{T}_i^*\} = \min_{\{\mathbf{T}_i\}} \!\sum_{(i, j) \in \mathcal{E}} \!\sum_{(u, v) \in \mathcal{M}_{i,j}} \!\!\rho \left( \| X_i^i(v) - \mathbf{T}_i \mathbf{T}_j^{-1} X_j^j(u) \|_2 \right),
%     \label{eq:pose_opt}
% \end{equation}
% where $\mathcal{K}$ denotes the set of keyframes, $X_j^j(u)$ and $X_i^i(v)$ represent the 3D point predictions for matched pixel pair $(u,v) \in \mathcal{M}_{i,j}$ in their local camera frames, and $\mathbf{T}_i \in \mathrm{SE}(3)$ denotes the absolute pose of keyframe $i$. The robust Huber loss $\rho(\cdot)$ effectively mitigates false correspondences caused by underwater optical degradation. \textbf{mapping frames} are selected when the average pixel parallax calculated over matched points in $\mathcal{M}_{t,k}$ with high geometric confidence $Q_t^t$ exceeds a baseline threshold $\tau_{\text{mapframe}}$, initiating the insertion of new 3D Gaussians into high-certainty areas; and \textbf{common frames}, which fail both criteria, are utilized solely to update and refine existing Gaussian parameters without expanding total map capacity.

Specifically, \textbf{keyframes} are spawned when the size of the reliable match set $|\mathcal{M}_{t,k}|$ falls below a predefined tracking threshold $\tau_{\text{keyframe}}$. This triggers the insertion of $\mathbf{I}_t$ into the backend pose graph to jointly optimize all absolute keyframe poses $\{\mathbf{T}_i\}_{i \in \mathcal{K}}$ across active spatial edges $(i,j) \in \mathcal{E}$ via:
\begin{equation}
    \{\mathbf{T}_i^*\} = \min_{\{\mathbf{T}_i\}} \!\sum_{(i, j) \in \mathcal{E}} \!\sum_{(u, v) \in \mathcal{M}_{i,j}} \!\!\rho \left( \| \mathbf{X}_i^i(v) - \mathbf{T}_i \mathbf{T}_j^{-1} \mathbf{X}_j^j(u) \|_2 \right),
    \label{eq:pose_opt}
\end{equation}
where $\mathcal{K}$ denotes the set of keyframes, $\mathbf{X}_j^j(u)$ and $\mathbf{X}_i^i(v)$ represent the 3D point predictions for matched pixel pair $(u,v) \in \mathcal{M}_{i,j}$ in their local camera frames, and $\mathbf{T}_i \in \mathrm{SE}(3)$ denotes the absolute pose of keyframe $i$. The robust Huber loss $\rho(\cdot)$ effectively mitigates false correspondences caused by underwater optical degradation. \textbf{Mapping frames} are selected when the average pixel parallax calculated over matched points in $\mathcal{M}_{t,k}$ with high geometric confidence $\mathbf{Q}_t^t$ exceeds a baseline threshold $\tau_{\text{mapframe}}$, initiating the insertion of new 3D Gaussians into high-certainty areas; and \textbf{common frames}, which fail both criteria, are utilized solely to update and refine existing Gaussian parameters.

% =========================================================
% 3.3 Hybrid Scene Representation
% =========================================================
\subsection{Hybrid Scene Representation}
\label{sec:representation}
\subsubsection{Distance-aware Structured Neural Gaussian}

% Most conventional underwater 3DGS-based reconstruction methods represent the scene using standard Gaussian primitives whose attributes are directly optimized. 
% Standard 3DGS mainly models appearance as a view-dependent property, while underwater observations exhibit strong distance-dependent color degradation due to attenuation and scattering. This makes conventional Gaussian primitives insufficient for representing underwater appearance variations caused by attenuation and scattering. To address this issue, we adopt a neural-enhanced Gaussian representation. For each Gaussian, we explicitly describe its observation condition using both the viewing direction and the observation distance.

In our framework, for each Gaussian primitive $\mathcal{G}_j$ in the dense map $\{\mathcal{G}_j\}_{j=1}^{M}$, its position $\mathbf{\mu}_j$ and opacity $\alpha_j$ are explicitly learned, whereas its structural attributes (rotation $q_j$ and scale $s_j$) along with the $0$-th order SH component $\mathbf{c}_{0,j}$ are dynamically decoded via a structure-aware, distance-guided perception decoder. 
This design bypasses the limitation of standard 3DGS, whose sparse, low-degree SH representations fail to capture complex, distance-dependent underwater color degradation. 
Specifically, the perception decoder processes structured Gaussian tokens to decode these attributes as residual increments relative to their initial states, effectively compensating for the coarse initialization inherent in streaming reconstruction. 
Consequently, each Gaussian primitive explicitly models appearance and geometry under both viewing direction and observation distance, enabling robust continuous scene modeling.

% Let \(\mathbf{p}_{cam}\) denote the camera center and \(\boldsymbol{\mu}_i\) the center of the \(i\)-th Gaussian.
% We define its observation distance and viewing direction as
% \begin{equation}
% d_i =
% \left\|
% \boldsymbol{\mu}_i - \mathbf{p}_{cam}
% \right\|_2,
% \qquad
% \mathbf{v}_i =
% \frac{
% \boldsymbol{\mu}_i - \mathbf{p}_{cam}
% }{
% d_i
% }.
% \end{equation}
% Here, \(d_i\) approximates the light propagation distance in the water medium, while \(\mathbf{v}_i\) describes the observation direction.
% They are concatenated into an observation condition:
% \begin{equation}
% \mathbf{o}_i = [\mathbf{v}_i, d_i] \in \mathbb{R}^{4}.
% \end{equation}

Concretely, let $\mathbf{p}_{\text{cam}}$ denote the camera center and $\boldsymbol{\mu}_i$ the center of the $i$-th Gaussian.
We define its observation distance and viewing direction as
\begin{equation}
d_i = \left\| \boldsymbol{\mu}_i - \mathbf{p}_{\text{cam}} \right\|_2, \qquad \mathbf{v}_i = \frac{\boldsymbol{\mu}_i - \mathbf{p}_{\text{cam}}}{d_i}.
\end{equation}
Here, $d_i$ approximates the light propagation distance in the water medium, while $\mathbf{v}_i$ describes the observation direction.
They are concatenated into an observation condition:
\begin{equation}
\mathbf{o}_i = [\mathbf{v}_i, d_i]^\top \in \mathbb{R}^{4}.
\end{equation}

% To model nonlinear observation-dependent appearance variations, we apply sinusoidal encoding to \(\mathbf{o}_i\):
% \begin{equation}
% \gamma(\mathbf{o}_i) =
% \left[
% \mathbf{o}_i,
% \left\{
% \sin(2^l \pi \mathbf{o}_i),
% \cos(2^l \pi \mathbf{o}_i)
% \right\}_{l=0}^{L-1}
% \right],
% \label{eq:obs_encoding}
% \end{equation}
% where \(L\) is the number of frequency bands.
% This encoding enables the representation to implicitly capture complex, range-dependent underwater imaging phenomena that are otherwise difficult to model explicitly.

To model nonlinear observation-dependent appearance variations, we apply sinusoidal encoding to $\mathbf{o}_i$:
\begin{equation}
\boldsymbol{\gamma}(\mathbf{o}_i) = \left[ \mathbf{o}_i, \left\{ \sin(2^l \pi \mathbf{o}_i), \cos(2^l \pi \mathbf{o}_i) \right\}_{l=0}^{L-1} \right],
\label{eq:obs_encoding}
\end{equation}
where $L$ is the number of frequency bands.
This encoding enables the representation to implicitly capture complex, range-dependent underwater imaging phenomena that are otherwise difficult to model explicitly.

% To maintain a stable Gaussian map and faithfully preserve continuous underwater terrains—such as sandy beds and mounds—we further enforce local structural organization during streaming reconstruction.
% Gaussians are voxelized by their 3D centers and grouped within each voxel, with each group containing at most \(K\) primitives.
% Each Gaussian has a learnable local feature \(\mathbf{f}^{l}_{i}\), while all Gaussians in the same group share a global feature \(\mathbf{f}^{g}_{g(i)}\) that encodes structural context.
% Together with the encoded observation condition \(\gamma(\mathbf{o}_i)\), the input feature of the \(i\)-th Gaussian is defined as
% \begin{equation}
% \mathbf{h}_i =
% \left[
% \mathbf{f}^{l}_{i},\ 
% \mathbf{f}^{g}_{g(i)},\ 
% \gamma(\mathbf{o}_i)
% \right],
% \end{equation}
% where \(g(i)\) denotes the group index.

To maintain a stable Gaussian map and faithfully preserve continuous underwater terrains, such as sandy beds and mounds, we further enforce local structural organization during streaming reconstruction.
Gaussians are incremently voxelized by their 3D centers and grouped within each voxel, with each group containing at most $K$ primitives.
Each Gaussian has a learnable local feature $\mathbf{f}_i^{\mathrm{l}}$, while all Gaussians in the same group share a global feature $\mathbf{f}_{g(i)}^{\mathrm{g}}$ that encodes structural context.
Together with the encoded observation condition $\boldsymbol{\gamma}(\mathbf{o}_i)$, the input feature of the $i$-th Gaussian is defined as
\begin{equation}
\mathbf{h}_i = \left[ \mathbf{f}_i^{\mathrm{l}},\, \mathbf{f}_{g(i)}^{\mathrm{g}},\, \boldsymbol{\gamma}(\mathbf{o}_i) \right]^\top,
\end{equation}
where $g(i)$ denotes the group index.

% We predict Gaussian parameters with a lightweight Local Gaussian Parameter Decoder (LGPD), which operates on each voxelized Gaussian group, as shown in Fig.~\ref{fig:pipeline}.
% Given at most \(K\) Gaussians in a group, LGPD first projects their input features into token embeddings:
% \begin{equation}
% \mathbf{e}_i = \mathrm{Linear}(\mathbf{h}_i).
% \end{equation}
% The token sequence is processed by a lightweight learnable neural network with a validity mask $\mathbf{m}_{1:K}$ for padded entries:
% \begin{equation}
% \mathbf{z}_{1:K}
% =
% \mathrm{Self\text{-}attention}
% \left(
% \mathbf{e}_{1:K},
% \mathbf{m}_{1:K}
% \right),
% \end{equation}
% which enables local context aggregation among neighboring Gaussians, thereby significantly enhancing the network's capacity to model non-linear underwater medium effects.

We predict Gaussian parameters with a lightweight Local Gaussian Parameter Decoder (LGPD), which operates on each voxelized Gaussian group, as shown in Fig.~\ref{fig:pipeline}.
Given at most $K$ Gaussians in a group, LGPD first projects their input features into token embeddings:
\begin{equation}
\mathbf{e}_i = \mathrm{Linear}(\mathbf{h}_i).
\end{equation}
The token sequence is processed by a lightweight learnable neural network with a validity mask $\mathbf{m}_{1:K}$ for padded entries:
\begin{equation}
\mathbf{z}_{1:K} = \mathrm{Self\text{-}Attention}\left(\mathbf{e}_{1:K}, \mathbf{m}_{1:K}\right),
\end{equation}
which enables local context aggregation among neighboring Gaussians, thereby significantly enhancing the network's capacity to model non-linear underwater medium effects.

% Each output token is then decoded into residual geometry and color updates:
% \begin{equation}
% [\Delta \mathbf{r}_i, \Delta \mathbf{s}_i]
% =
% \mathrm{MLP}_{geo}(\mathbf{z}_i),
% \qquad
% \Delta \mathbf{c}_i
% =
% \mathrm{MLP}_{color}(\mathbf{z}_i).
% \end{equation}
% Here, \(\Delta \mathbf{r}_i\), \(\Delta \mathbf{s}_i\), and \(\Delta \mathbf{c}_i\) are applied as residual corrections to the initialized Gaussian parameters, producing the updated rotation, scale, and color of the \(i\)-th Gaussian.
% This residual formulation improves the stability of streaming optimization.

Each output token is then decoded into residual geometry and color updates:
\begin{equation}
[\Delta \mathbf{r}_i, \Delta \mathbf{s}_i] = \mathrm{MLP}_{\mathrm{geo}}(\mathbf{z}_i), \qquad \Delta \mathbf{c}_i = \mathrm{MLP}_{\mathrm{color}}(\mathbf{z}_i).
\end{equation}
Here, $\Delta \mathbf{r}_i$, $\Delta \mathbf{s}_i$, and $\Delta \mathbf{c}_i$ are applied as residual corrections to the initialized Gaussian parameters, producing the updated rotation, scale, and color of the $i$-th Gaussian.
This residual formulation improves the stability of streaming optimization.

\subsubsection{Underwater Medium Modeling}

% Underwater image formation is mainly governed by light attenuation and backscattering. Inspired by the simplified image formation model in \cite{akkaynak2018revised}, we model the observed color as the combination of attenuated scene radiance and accumulated backscatter:
% \begin{equation}
% I = J \cdot \exp(-\mathbf{B}_a z) + B(z),
% \end{equation}
% where \(J\) denotes the medium-free scene radiance rendered from the Gaussian map, \(z\) denotes the depth along the camera ray estimated from the Gaussian-rendered depth, \(\mathbf{B}_a\) is the RGB attenuation coefficient, and \(B(z)\) represents the backscatter term.

Underwater image formation is mainly governed by light attenuation and backscattering. Inspired by the simplified image formation model in \cite{akkaynak2018revised}, we model the observed color as the combination of attenuated scene radiance and accumulated backscatter:
\begin{equation}
\mathbf{I} = \mathbf{J} \cdot \exp(-\boldsymbol{\beta}_{\mathrm{a}} z) + \mathbf{B}(z),
\label{eq:underwater_formation}
\end{equation}
where $\mathbf{J}$ denotes the medium-free scene radiance rendered from the Gaussian map, $z$ denotes the depth along the camera ray estimated from the Gaussian-rendered depth, $\boldsymbol{\beta}_{\mathrm{a}}$ is the RGB attenuation coefficient vector, and $\mathbf{B}(z)$ represents the backscatter term.

% To enhance the representation capability of our model, we assume that the medium parameters are directional-dependent rather than isotropic. Consequently, we predict the attenuation coefficient conditioned directly on the camera orientation:
% Given the camera rotation \(\mathbf{R}_t \in \mathbb{R}^{3 \times 3}\) of frame \(t\), we flatten it into a pose vector:
% \begin{equation}
% \mathbf{p}_t = \mathrm{vec}(\mathbf{R}_t) \in \mathbb{R}^{9}.
% \end{equation}
% Using the same sinusoidal encoding as in Eq.~(\ref{eq:obs_encoding}), we obtain the pose embedding \(\gamma(\mathbf{p}_t)\), and decode the frame-specific attenuation coefficient as
% \begin{equation}
% \mathbf{B}^{t}_a = \mathrm{MLP}_{a}(\gamma(\mathbf{p}_t)).
% \end{equation}

To enhance the representation capability of our model, we assume that the medium parameters are directional-dependent rather than isotropic. Consequently, we predict the attenuation coefficient conditioned directly on the camera orientation:
Given the camera rotation $\mathbf{R}_t \in \mathbb{R}^{3 \times 3}$ of frame $t$, we flatten it into a pose vector:
\begin{equation}
\mathbf{p}_t = \mathrm{vec}(\mathbf{R}_t) \in \mathbb{R}^{9}.
\end{equation}
Using the same sinusoidal encoding as in Eq.~(\ref{eq:obs_encoding}), we obtain the pose embedding $\boldsymbol{\gamma}(\mathbf{p}_t)$, and decode the frame-specific attenuation coefficient as
\begin{equation}
\boldsymbol{\beta}_{\mathrm{a}}^t = \mathrm{MLP}_{\mathrm{a}}(\boldsymbol{\gamma}(\mathbf{p}_t)).
\end{equation}

% For backscatter, we adopt a physics-inspired RGB formulation:
% \begin{equation}
% B_c(z) =
% B^{\infty}_c
% \left(1 - \exp(-b_c z)\right)
% +
% B^{res}_c \exp(-d_c z)
% \end{equation}
% Here, \(B^{\infty}_c\) denotes the asymptotic backscatter intensity at infinite distance, \(b_c\) controls the backscatter accumulation rate, \(B^{res}_c\) models the near-distance residual backscatter, and \(d_c\) is the channel-wise residual decay parameter. The parameters \(\{B^{\infty}_c, b_c, B^{res}_c, d_c\}_{c\in\{r,g,b\}}\) are learnable channel-wise parameters, resulting in 12 learnable parameters for RGB backscatter modeling.

For backscatter, we adopt a physics-inspired RGB formulation:
\begin{equation}
\mathbf{B}(z) = \mathbf{B}_{\infty} \odot \left(\mathbf{1} - \exp(-\mathbf{b} z)\right) + \mathbf{B}_{\mathrm{res}} \odot \exp(-\mathbf{d} z),
\label{eq:backscatter_model}
\end{equation}
where $\mathbf{B}_{\infty} \in \mathbb{R}^3$ denotes the asymptotic backscatter intensity vector at infinite distance, $\mathbf{b} \in \mathbb{R}^3$ controls the channel-wise backscatter accumulation rate, $\mathbf{B}_{\mathrm{res}} \in \mathbb{R}^3$ models the near-distance residual backscatter vector, and $\mathbf{d} \in \mathbb{R}^3$ represents the channel-wise residual decay parameter vector. The parameters $\{\mathbf{B}_{\infty}, \mathbf{b}, \mathbf{B}_{\mathrm{res}}, \mathbf{d}\}$ are learnable channel-wise vectors.

\subsubsection{Medium-guided Incremental Gaussian Initialization}
For keyframes and mapping frames selected by the frontend, as described in Section~\ref{sec:pose}, new Gaussians are inserted to incorporate newly observed scene regions.
Instead of initializing them directly from degraded underwater images, we use the online learning medium model to obtain medium-reduced observations.

% Given an underwater image \(\mathbf{I}_t\), the camera pose estimated by the frontend, and the medium parameters accumulated from previous mapping steps, we estimate the attenuation coefficient \(\mathbf{B}_a^t\) and the backscatter term \(\mathbf{B}(z)\).
% The medium-reduced image is recovered as
% \begin{equation}
% \hat{\mathbf{J}}_t
% =
% \left(\mathbf{I}_t - \mathbf{B}(z)\right)
% \odot
% \exp\left(\mathbf{B}_a^t z\right),
% \end{equation}
% where \(z\) denotes the depth along the camera ray rendered from the current Gaussian map, and \(\odot\) denotes element-wise multiplication.
% Compared with the raw underwater observation, \(\hat{\mathbf{J}}_t\) provides cleaner appearance cues for Gaussian initialization.

Given an underwater image $\mathbf{I}_t$, the camera pose estimated by the frontend, and the medium parameters accumulated from previous mapping steps, we estimate the attenuation coefficient $\boldsymbol{\beta}_{\mathrm{a}}^t$ and the backscatter term $\mathbf{B}(z)$.
The medium-reduced image is recovered as
\begin{equation}
\hat{\mathbf{J}}_t = \left(\mathbf{I}_t - \mathbf{B}(z)\right) \odot \exp\left(\boldsymbol{\beta}_{\mathrm{a}}^t z\right),
\end{equation}
where $z$ denotes the depth along the camera ray rendered from the current Gaussian map, and $\odot$ denotes element-wise multiplication.
Compared with the raw underwater observation, $\hat{\mathbf{J}}_t$ provides cleaner appearance cues for Gaussian initialization.

% To determine where new Gaussians should be inserted, we compare the medium-reduced observation \(\hat{\mathbf{J}}_t\) with the medium-free rendering \(\tilde{\mathbf{J}}_t\) from the current Gaussian map.
% We compute a structure-aware insertion probability using a Laplacian-of-Gaussian operator:
% \begin{equation}
% \begin{aligned}
% P_a(u,v) =
% \max\Big(
% & \min\big(
% \left\| \nabla^2 G_{\sigma} * \hat{\mathbf{J}}_t(u,v) \right\|, 1
% \big) \\
% & -
% \min\big(
% \left\| \nabla^2 G_{\sigma} * \tilde{\mathbf{J}}_t(u,v) \right\|, 1
% \big),
% 0
% \Big),
% \end{aligned}
% \end{equation}
% where \(G_{\sigma}\) is a Gaussian kernel with standard deviation \(\sigma\), \(\nabla^2\) is the Laplacian operator, and \(*\) denotes convolution.
% Pixels with \(P_a(u,v) > \tau_a\) are selected for Gaussian insertion. For each newly inserted Gaussian, the base color is initialized from the medium-reduced image \(\hat{\mathbf{J}}_t\) rather than the raw underwater image \(\mathbf{I}_t\).
% This base color is kept fixed, while the neural-enhanced Gaussian representation predicts residual color updates during subsequent optimization.

To determine where new Gaussians should be inserted, we compare the medium-reduced observation $\hat{\mathbf{J}}_t$ with the medium-free rendering $\tilde{\mathbf{J}}_t$ from the current Gaussian map.
We compute a structure-aware insertion probability using a Laplacian-of-Gaussian operator:
\begin{equation}
\begin{aligned}
P_{\mathrm{a}}(u,v) = \max\Big( & \min\big( \left\| \nabla^2 G_{\sigma} * \hat{\mathbf{J}}_t(u,v) \right\|_2, 1 \big) \\
& - \min\big( \left\| \nabla^2 G_{\sigma} * \tilde{\mathbf{J}}_t(u,v) \right\|_2, 1 \big), 0 \Big),
\end{aligned}
\end{equation}
where $G_{\sigma}$ is a Gaussian kernel with standard deviation $\sigma$, $\nabla^2$ is the Laplacian operator, and $*$ denotes convolution.
Pixels with $P_{\mathrm{a}}(u,v) > \tau_{\mathrm{a}}$ are selected for Gaussian insertion. For each newly inserted Gaussian, the base color is initialized from the medium-reduced image $\hat{\mathbf{J}}_t$ rather than the raw underwater image $\mathbf{I}_t$.
This base color is kept fixed, while the neural-enhanced Gaussian representation predicts residual color updates during subsequent optimization.

\subsection{Loss Function}
\label{loss_function}
After initialization, we jointly optimize the Gaussian attributes, local and global Gaussian features, LGPD parameters, and underwater medium parameters through differentiable rendering. 

Specifically, to compensate for boundary lens distortions and prioritize central view fidelity, a radial decay kernel $\mathbf{K}_{\mathrm{rad}} \in \mathbb{R}^{H \times W}$ weights the spatial loss terms. For common frames, we construct an adaptive binary validity mask $\mathbf{M} \in \{0, 1\}^{H \times W}$ based on the weighted residual to filter out outliers:
\begin{equation}
\mathbf{M}(u,v) = \begin{cases} 
0, & \text{if } \exists \, c \in \{R,G,B\}, \\
   & \mathbf{K}_{\mathrm{rad}}(u,v) \cdot |\hat{\mathbf{I}}_c(u,v) - \mathbf{I}_c(u,v)| > \tau_e, \\
1, & \text{otherwise,}
\end{cases}
\end{equation}
where $\hat{\mathbf{I}}$ and $\mathbf{I}$ are the rendered and ground-truth images. For keyframes and mapping frames, $\mathbf{M}$ is set to an all-ones matrix.

Using $\mathbf{M}$, we compute the masked rendered image $\hat{\mathbf{I}}_{\mathbf{M}} = \mathbf{M} \odot \hat{\mathbf{I}}$, ground-truth image $\mathbf{I}_{\mathbf{M}} = \mathbf{M} \odot \mathbf{I}$, rendered inverse depth $\hat{\mathbf{D}}_{\mathbf{M}} = \mathbf{M} \odot \hat{\mathbf{D}}$, and masked monocular depth prior $\mathbf{D}_{\mathbf{M}} = \mathbf{M} \odot \mathbf{D}$, where the raw depth prior $\mathbf{D}$ is provided by our underwater 3D vision foundation model. The total optimization objective $\mathcal{L}_{\mathrm{total}}$ combines the color, depth, and regularization terms:
\begin{equation}
\begin{aligned}
\mathcal{L}_{\mathrm{total}} = \; & (1 - \lambda_{\mathrm{ssim}}) \mathcal{L}_{1} + \lambda_{\mathrm{ssim}} \mathcal{L}_{\mathrm{ssim}} \\
& + \lambda_{d} \mathcal{L}_{\mathrm{depth}} + \lambda_{s} \mathcal{L}_{\mathrm{scale}},
\end{aligned}
\label{eq:total_loss}
\end{equation}
where the individual loss terms are defined as:
\begin{align}
\mathcal{L}_{1} &= \frac{\sum_{u,v} \mathbf{K}_{\mathrm{rad}}(u,v) \cdot \left| \hat{\mathbf{I}}_{\mathbf{M}}(u,v) - \mathbf{I}_{\mathbf{M}}(u,v) \right|}{HW} , \\
\mathcal{L}_{\mathrm{depth}} &= \frac{\sum_{u,v} \mathbf{K}_{\mathrm{rad}}(u,v) \cdot \left| \hat{\mathbf{D}}_{\mathbf{M}}(u,v) - \mathbf{D}_{\mathbf{M}}(u,v) \right|}{HW}, \\
\mathcal{L}_{\mathrm{ssim}} &= 1 - \mathrm{SSIM}(\hat{\mathbf{I}}_{\mathbf{M}}, \mathbf{I}_{\mathbf{M}}), \\
\mathcal{L}_{\mathrm{scale}} &= \frac{1}{N} \sum_{i=1}^{N} \prod_{j=1}^{3} s_{i,j}.
\end{align}
Here, $\lambda_{\mathrm{ssim}}$, $\lambda_{d}$, and $\lambda_{s}$ denote the balancing weights for structural similarity, depth consistency, and Gaussian scale regularization across $N$ Gaussians, respectively.

\begin{figure*}[t]
\centering
\includegraphics[width=\textwidth]{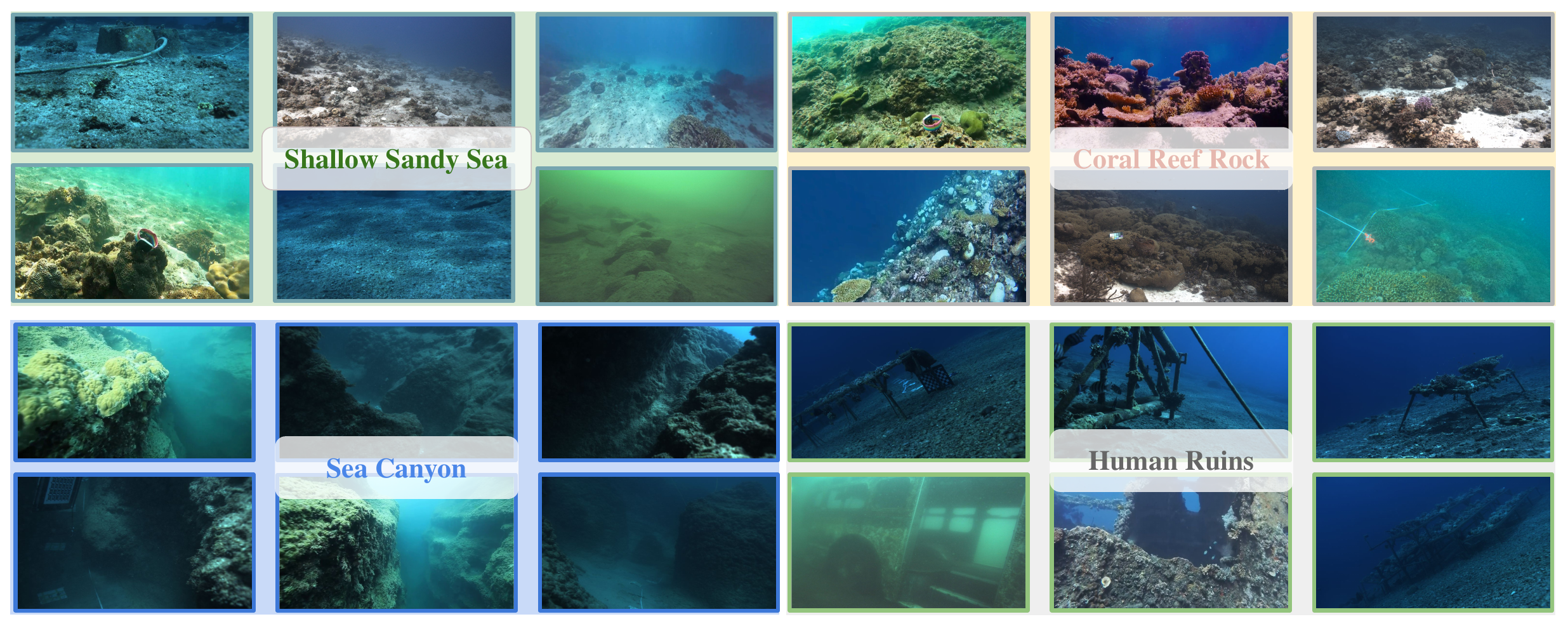}
\caption{\textbf{Overview and taxonomy of our proposed underwater evaluation benchmark.} The dataset spans broad physical spatial scales and rich topological diversity, encompassing characteristic benthic environments such as shallow sandy beds, canyons, coral reefs, rocky terrains, submerged caves, and artificial ruins. Furthermore, it incorporates diverse camera trajectories (e.g., straight paths, loops, and erratic 3D motions) across a wide range of frame lengths: 14 short sequences ($0\text{--}50$ frames), 13 medium-short sequences ($50\text{--}100$ frames), 20 medium sequences ($100\text{--}200$ frames), 3 long sequences ($200\text{--}400$ frames), and 12 extended-range sequences ($>400$ frames) to comprehensively assess online tracking and long-term mapping stability.}
\label{fig:diverse_test_data}
\end{figure*}

\begin{table}[t]
\centering
\footnotesize
\caption{\textbf{Composition of underwater training image pairs.} The datasets cover diverse synthetic and real underwater scenes.}
\label{tab:training_pairs}
\setlength{\tabcolsep}{1.7pt}
\renewcommand{\arraystretch}{1.1}
\begin{tabularx}{\linewidth}{@{} l r c X @{}}
\toprule
Dataset & Pairs & Type & Scene Content \\
\midrule
TartanAir-Ocean & 76,387  & Syn.  & Seafloor, terrain, industrial, wrecks \\
MIMIR-UW        & 44,721  & Syn.  & Seafloor, algae, industrial, ruins \\
UWStereo        & 26,611  & Syn.  & Corals, vessels, structures, ROVs \\
FLSea-stereo    & 6,576   & Real  & Seafloor, canyons, corals, artifacts \\
sweet-corals    & 69,978  & Real  & Corals, complex biomes \& structures \\
\midrule
Total           & 224,273 & Mixed & Varied benthic biomes and subsea structures \\
\bottomrule
\end{tabularx}
\end{table}

\begin{table*}[t]
\centering
\small
\caption{\textbf{Quantitative reconstruction results on real-world underwater datasets.} We report PSNR, SSIM, and LPIPS. Higher PSNR/SSIM and lower LPIPS indicate better performance. The best and second-best results are in \textbf{bold} and \underline{underline}.}
\label{tab:reconstruction_results}
\setlength{\tabcolsep}{1.5pt}
\renewcommand{\arraystretch}{1.1}
\resizebox{\textwidth}{!}{
\begin{tabular}{lcccccccccccccccccc}
\toprule
\multirow{2}{*}{Method} 
& \multicolumn{3}{c}{Canyons} 
& \multicolumn{3}{c}{RedSea} 
& \multicolumn{3}{c}{UW-Stereo-VI} 
& \multicolumn{3}{c}{SeaThru-NeRF} 
& \multicolumn{3}{c}{S-UW} 
& \multicolumn{3}{c}{Internet Data} \\
\cmidrule(lr){2-4}
\cmidrule(lr){5-7}
\cmidrule(lr){8-10}
\cmidrule(lr){11-13}
\cmidrule(lr){14-16}
\cmidrule(lr){17-19}
& PSNR$\uparrow$ & SSIM$\uparrow$ & LPIPS$\downarrow$
& PSNR$\uparrow$ & SSIM$\uparrow$ & LPIPS$\downarrow$
& PSNR$\uparrow$ & SSIM$\uparrow$ & LPIPS$\downarrow$   
& PSNR$\uparrow$ & SSIM$\uparrow$ & LPIPS$\downarrow$   
& PSNR$\uparrow$ & SSIM$\uparrow$ & LPIPS$\downarrow$
& PSNR$\uparrow$ & SSIM$\uparrow$ & LPIPS$\downarrow$ \\
\midrule
UW-GS           & 27.33          & 0.916          & 0.288          & 23.78          & \underline{0.809} & 0.316          & 27.87          & 0.903          & 0.366          & \textbf{25.79} & \textbf{0.858} & \textbf{0.260} & 20.55          & \underline{0.733} & \underline{0.279} & 23.56          & 0.786          & 0.311 \\
OnTheFly-NVS    & 27.20          & 0.894          & 0.317          & 22.37          & 0.699             & 0.387          & 29.32          & 0.888          & 0.385          & 22.12          & 0.719          & \underline{0.324} & 19.51          & 0.623             & 0.335          & 23.42          & 0.746          & 0.339 \\
HI-SLAM2        & 29.88          & 0.898          & 0.269          & 23.66          & 0.716             & 0.448          & 27.69          & 0.856          & 0.423          & \underline{25.33} & \underline{0.745} & 0.471          & 21.42          & 0.685             & 0.367          & 25.13          & 0.769          & 0.369 \\
MonoGS          & 27.83          & 0.875          & 0.460          & 20.89          & 0.644             & 0.708          & 25.87          & 0.861          & 0.480          & 13.82          & 0.524          & 0.589          & 9.70           & 0.261             & 0.667          & 16.47          & 0.607          & 0.623 \\
S3PO-GS         & 29.62          & 0.910          & 0.369          & 22.47          & 0.669             & 0.562          & 29.96          & 0.900          & 0.450          & 19.48          & 0.609          & 0.493          & 19.48          & 0.569             & 0.459          & 22.82          & 0.743          & 0.491 \\
ARTDECO         & \underline{31.83} & \underline{0.933} & \underline{0.262} & \underline{25.41} & 0.808      & \underline{0.300} & \underline{36.22} & \underline{0.913} & \underline{0.337} & 23.75          & 0.689          & 0.378          & \underline{23.05} & 0.703             & 0.302          & \underline{26.96} & \underline{0.821} & \underline{0.279} \\
WaterSplat-SLAM & 29.33          & 0.885          & 0.303          & 22.61          & 0.726             & 0.401          & 27.06          & 0.896          & 0.432          & 24.43          & 0.736          & 0.492          & 19.59          & 0.605             & 0.539          & 23.22          & 0.747          & 0.355 \\
Ours        & \textbf{34.53} & \textbf{0.955} & \textbf{0.199} & \textbf{26.77} & \textbf{0.836}    & \textbf{0.272} & \textbf{36.31} & \textbf{0.920} & \textbf{0.303} & 24.83          & 0.739          & 0.346          & \textbf{23.73} & \textbf{0.741}    & \textbf{0.277} & \textbf{28.53} & \textbf{0.852} & \textbf{0.241} \\
\bottomrule
\end{tabular}
}
\end{table*}

\begin{figure*}[t]
\centering
\includegraphics[width=\textwidth]{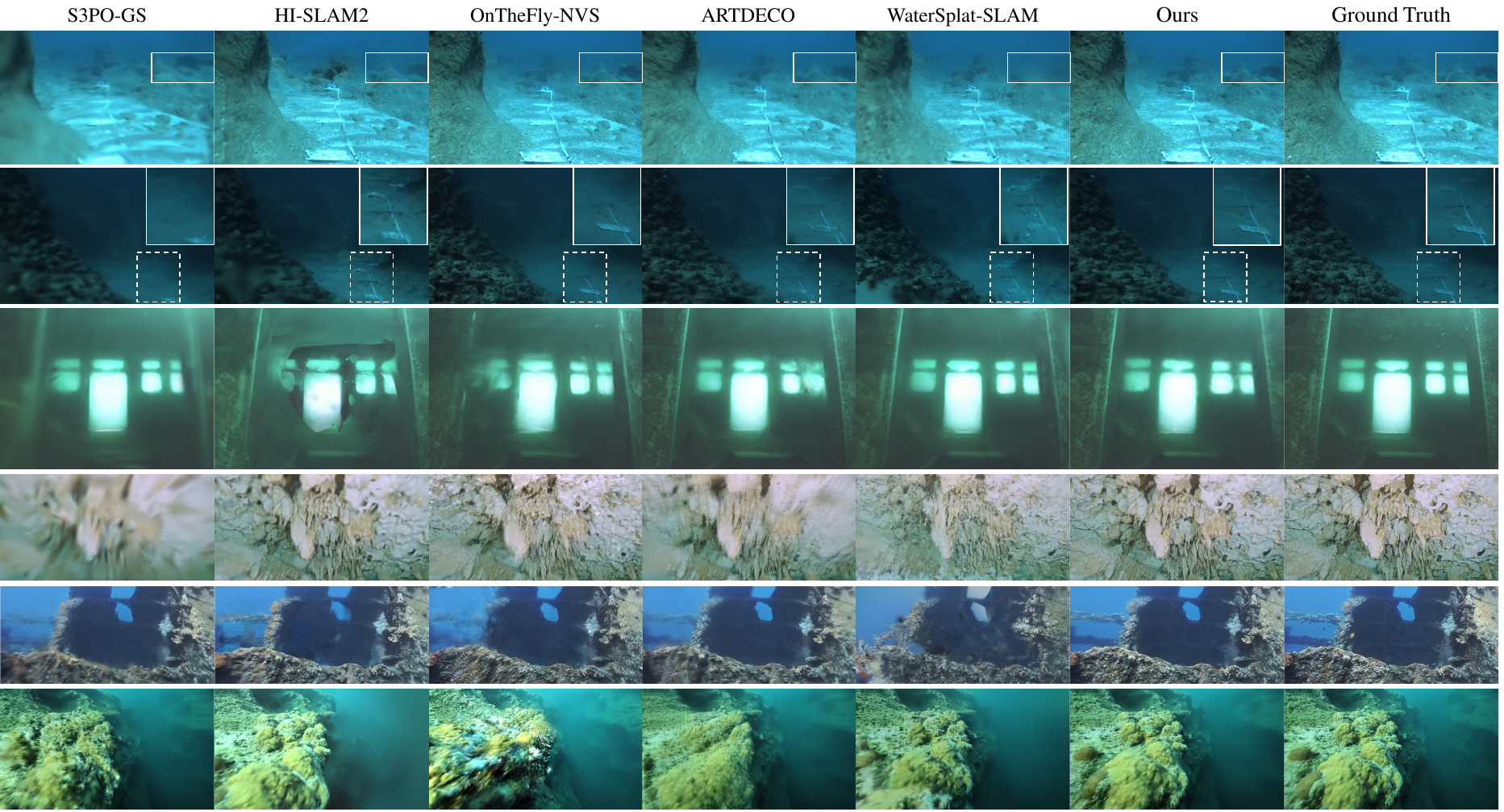}
\caption{\textbf{Qualitative comparisons of AquaFlow against baselines~\cite{cheng2025outdoor,zhang2025hi,meuleman2025fly,li2026artdeco,wang2026watersplat}.} 
The top three rows highlight AquaFlow’s superior modeling capability for distant underwater scenes and intricate seafloor details (e.g., ground markings and discarded vehicle textures). 
The bottom three rows demonstrate our strength in recovering complex subsea structural topologies (e.g., underwater caves, shipwreck ruins and canyons), where domain-adapted 3D geometric priors and appearance representations effectively preserve sharp object boundaries and detailed structures compared to baseline methods.}
\label{fig:rendering}
\end{figure*}

\section{Experiment}
\subsection{Experiment Setup}
\subsubsection{Test Datasets \& Evaluation Metrics}
To evaluate streaming underwater reconstruction under open-world conditions, we assemble a comprehensive 62-sequence test suite, combining 25 sequences from curated public benchmarks with 37 in-the-wild internet sequences. 
As illustrated in Fig.~\ref{fig:diverse_test_data}, our test benchmark exhibits high diversity across visual scenes, physical scales, and trajectory lengths. 

Specifically, the public subset spans both small- and large-scale underwater scenarios: 
the small-scale domain includes 4 sequences from SeaThru-NeRF~\cite{levy2023seathru} and 4 sequences from S-UW~\cite{wang2025uw}, while the large-scale domain comprises 4 sequences from \textit{Canyons} and 8 sequences from \textit{RedSea} (both sourced from FLSea~\cite{randall2023flsea}), alongside 5 sequences from UW-Stereo-VI~\cite{joshi2019experimental}. 
Collectively, these public sequences cover a broad spectrum of underwater scene types, including seafloor terrain, coral reefs, caves, submerged vehicles, and man-made structures. 
To further push system limits under unconstrained real-world conditions, the internet subset incorporates 37 sequences that extend this diversity across unconstrained scene categories, spatial scales, and trajectory lengths.
For these uncalibrated internet sequences, camera intrinsics and pseudo ground-truth camera poses are reconstructed via structure-from-motion (SfM) using COLMAP~\cite{schoenberger2016sfm}, accompanied by meticulous manual inspection of point-cloud geometric consistency to guarantee evaluation fidelity.

Reconstruction quality is quantitatively assessed using Peak Signal-to-Noise Ratio (PSNR), Structural Similarity Index (SSIM)~\cite{wang2004image}, and Learned Perceptual Image Patch Similarity (LPIPS)~\cite{zhang2018unreasonable}. 
Camera tracking performance is evaluated via the Root Mean Square Error (RMSE) of Absolute Trajectory Error (ATE) after rigid trajectory alignment.
In addition, we measure reconstruction time and per-frame processing latency to rigorously compare the computational overhead against both streaming and off-line baselines.

\subsubsection{Baselines}
We benchmark AquaFlow against state-of-the-art methods across multiple tracking, rendering, and efficiency settings in challenging underwater environments. 
For camera pose estimation, we compare localization accuracy against optical-flow-based DROID-SLAM~\cite{teed2021droidslam}, vision foundation model-based SLAMs (MASt3R-SLAM~\cite{murai2024mast3r}, VGGT-SLAM~\cite{maggio2026vggt}, and VGGT-SLAM2~\cite{maggio2026vggt2}), as well as UFEN-SLAM~\cite{yang2023knowledge} which is a domain-specific underwater SLAM framework that replaces classical ORB feature extraction with learned underwater-adapted features. 
For rendering quality, we evaluate against streaming novel view synthesis frameworks, including MonoGS~\cite{matsuki2024gaussian}, S3PO-GS~\cite{cheng2025outdoor}, OnTheFly-NVS~\cite{meuleman2025fly}, and HI-SLAM2~\cite{zhang2025hi}. 
We also include UW-GS~\cite{levy2023seathru}, an offline underwater Gaussian Splatting system, whose optimization iterations are carefully adjusted to ensure a fair and reasonable comparison between streaming and offline rendering capabilities. 
Furthermore, ARTDECO~\cite{li2026artdeco} and WaterSplat-SLAM~\cite{wang2026watersplat} serve as strong unified baselines, where both camera tracking accuracy and novel view reconstruction fidelity are benchmarked concurrently. 
Finally, computational runtime and streaming efficiency are quantitatively evaluated against UW-GS, ARTDECO, and WaterSplat-SLAM.

\subsubsection{3D Vision Foundation Model Training Data}
To adapt dense geometry prediction and 3D reconstruction vision foundation models~\cite{leroy2024grounding} to underwater environments, we construct a dedicated training dataset comprising 224,273 valid image pairs, curated and repurposed from five public underwater datasets (Tab.~\ref{tab:training_pairs}).
The selected sequences contain typical underwater visual degradations, such as color attenuation haze, scattering, low contrast, and texture loss, while maintaining sufficient viewpoint overlap and temporal continuity for reliable geometric supervision.
By strategically combining synthetic data (providing large-scale, consistent geometry) with real-world scenes (introducing natural medium degradation and realistic visual features), this dataset enhances MASt3R's geometric representation while drastically reducing domain shifts when deployed in real underwater conditions.

% \begin{table}[t]
% \centering
% \footnotesize
% \caption{\textbf{Composition of underwater training image pairs.} The datasets cover diverse synthetic and real underwater scenes.}
% \label{tab:training_pairs}
% \setlength{\tabcolsep}{2.5pt}
% \renewcommand{\arraystretch}{1.05}
% \begin{tabularx}{\linewidth}{l r c X}
% \toprule
% Dataset & Pairs & Type & Scene Content \\
% \midrule
% TartanAir-Ocean & 76,387 & Syn. & Seafloor, rocks, terrain, industrial structures, and shipwreck-like scenes. \\
% MIMIR-UW & 44,721 & Syn. & Seafloor, algae, inspection scenes, industrial scenes, and ruin-like scenes. \\
% UWStereo & 26,611 & Syn. & Coral reefs, ships, robots, and underwater structures. \\
% FLSea-stereo & 6,576 & Real & Seafloor, canyons, rocks, corals, and man-made objects. \\
% sweet-corals & 69,978 & Real & Coral reefs, complex structures, and natural textures. \\
% \midrule
% Total & 224,273 & Syn.+Real & Diverse terrain, coral reefs, underwater structures, and ruin-like scenes. \\
% \bottomrule
% \end{tabularx}
% \end{table}

\subsubsection{Implementation Details.}
% We fine-tune MASt3R on 16 NVIDIA RTX 4090 GPUs for 30 epochs with a per-GPU batch size of 2 and gradient accumulation over 4 iterations, yielding an effective batch size of 128. Optimization is performed using AdamW with $\boldsymbol{\beta}_{\mathrm{AdamW}} = (0.9, 0.95)$ and a loss weight of $\beta = 0.075$. We apply a 4-epoch linear warmup followed by a cosine learning rate decay schedule from $1 \times 10^{-6}$ down to $1 \times 10^{-7}$.

% All streaming reconstruction experiments and runtime evaluations are executed on a desktop workstation with an Intel Core i9-14900K CPU and a single NVIDIA RTX 4090 GPU. 
% Following standard novel view synthesis protocols, every 8th frame of each sequence is held out for evaluation and excluded from map construction, while its camera pose is still tracked for trajectory assessment. 
% For system hyperparameters, we set the voxel size to $0.2$\,m and assign 32 tokens per voxel group.

We fine-tune MASt3R on 16 NVIDIA RTX 4090 GPUs for 30 epochs with a per-GPU batch size of 2 and gradient accumulation over 4 iterations. Optimization is performed using AdamW with $\boldsymbol{\beta}_{\mathrm{AdamW}} = (0.9, 0.95)$ and a loss weight of $\beta = 0.075$. We apply a 4-epoch linear warmup followed by a cosine learning rate decay schedule from $1 \times 10^{-6}$ down to $1 \times 10^{-7}$.

All experiments are conducted on an Intel i9-14900K CPU and a single NVIDIA RTX 4090 GPU. Following standard protocols, every 8th frame is held out for evaluation, with its pose tracked for trajectory assessment. For implementation details, we set the voxel size to $0.2\,\text{m}$, the token capacity per voxel group to $N_{\text{token}} = 32$, and the positional encoding bandwidth to $L = 4$. Keyframe selection, mapping parallax, and Gaussian insertion thresholds are configured as $\tau_{\text{keyframe}} = 0.7$, $\tau_{\text{mapframe}} = 30\,\text{pixels}$, and $\tau_{\mathrm{a}} = 1.0$, respectively. For optimization, the loss weight for $\mathcal{L}_{\mathrm{l1}}$ is set to $0.8$. $\lambda_{\mathrm{ssim}} = 0.2$, $\lambda_{d} = 0.01$ for depth loss, and $\lambda_{s} = 0.01$ for scaling regularization.

\begin{table*}[t]
\centering
\small
\caption{\textbf{Quantitative efficiency and Gaussian number comparison.} We report scene training time ($\text{Time (s)}\downarrow$), training throughput ($\text{FPS}\uparrow$), and Gaussian count ($\text{\#GS (k)}\downarrow$). The best and second-best results are in \textbf{bold} and \underline{underline}, respectively.}
\label{tab:time_efficiency}
\setlength{\tabcolsep}{1.5pt}
\renewcommand{\arraystretch}{1.1}
\resizebox{\textwidth}{!}{ 
\begin{tabular}{lcccccccccccccccccc}
\toprule
\multirow{2}{*}{Method} 
& \multicolumn{3}{c}{FLsea\_canyons} 
& \multicolumn{3}{c}{FLsea\_redsea} 
& \multicolumn{3}{c}{underwater-stereo-vi} 
& \multicolumn{3}{c}{seathruNerf} 
& \multicolumn{3}{c}{S-UW} 
& \multicolumn{3}{c}{Internet data} \\
\cmidrule(lr){2-4} \cmidrule(lr){5-7} \cmidrule(lr){8-10} \cmidrule(lr){11-13} \cmidrule(lr){14-16} \cmidrule(lr){17-19}
& Time$\downarrow$ & FPS$\uparrow$ & \#GS(k)$\downarrow$
& Time$\downarrow$ & FPS$\uparrow$ & \#GS(k)$\downarrow$
& Time$\downarrow$ & FPS$\uparrow$ & \#GS(k)$\downarrow$
& Time$\downarrow$ & FPS$\uparrow$ & \#GS(k)$\downarrow$
& Time$\downarrow$ & FPS$\uparrow$ & \#GS(k)$\downarrow$
& Time$\downarrow$ & FPS$\uparrow$ & \#GS(k)$\downarrow$ \\
\midrule
UW-GS           & 1896.0 & 0.26          & 952.0            & 4174.0         & 0.12          & 3120.0           & 557.6          & 0.31          & 750.9           & 227.5         & 0.10          & 707.8           & 245.2         & 0.10          & 1032.0           & 557.0          & 0.06          & 1132.9 \\
ARTDECO         & \underline{498.4} & \underline{1.00} & \underline{464.2} & \textbf{593.3} & \textbf{0.84} & 1006.4           & \underline{212.8} & \underline{0.74} & 733.9          & \underline{50.6} & \underline{0.43} & 917.4          & \textbf{39.8} & \textbf{0.60} & 1214.2           & \textbf{115.0} & \textbf{0.92} & \textbf{705.2} \\
WaterSplat-SLAM & 1562.9 & 0.32          & \textbf{13.8}    & 1029.0         & 0.48          & \textbf{26.9}    & 757.8          & 0.21          & \textbf{7.4}    & 75.9          & 0.29          & \textbf{33.8}   & 88.5          & 0.27          & \textbf{37.1}    & \underline{147.6} & 0.71          & \underline{750.0} \\
Ours        & \textbf{467.1} & \textbf{1.08} & 610.5            & \underline{709.4} & \underline{0.69} & \underline{975.7} & \textbf{207.2} & \textbf{0.76} & \underline{581.9} & \textbf{38.5} & \textbf{0.56} & \underline{790.2} & \underline{47.3} & \underline{0.51} & \underline{1157.8} & 153.9          & \underline{0.68} & 758.6 \\
\bottomrule
\end{tabular}
}
\end{table*}

\begin{table}[t]
\centering
\small
\caption{\textbf{Quantitative localization (trajectory) results on underwater datasets.} We report ATE RMSE (m) $\downarrow$. The best and second-best results are in \textbf{bold} and \underline{underline}, respectively.}
\label{tab:tracking_results}
\setlength{\tabcolsep}{1.5pt}
\renewcommand{\arraystretch}{1.1}
\resizebox{\columnwidth}{!}{
\begin{tabular}{lcccccc}
\toprule
Method 
& Canyons 
& RedSea 
& UW-Stereo-VI 
& SeaThru-NeRF 
& S-UW 
& Internet \\
\midrule
UFEN-SLAM       & 1.107 & 1.226 & 0.978 & 1.637 & 1.570 & 1.477 \\
DROID-SLAM      & 0.935 & \textbf{0.737} & 2.969 & 0.376 & \textbf{0.069} & 1.057 \\
MASt3r-SLAM     & 0.978 & 1.357 & 0.872 & 1.579 & 1.270 & 0.997 \\
VGGT-SLAM       & 3.079 & 2.747 & 1.965 & 0.237 & 0.469 & 2.261 \\
VGGT-SLAM2      & 0.732 & 1.206 & 0.520 & 0.240 & 0.470 & 1.339 \\
ARTDECO         & 0.559 & 0.991 & 0.325 & 0.372 & 0.245 & \underline{0.593} \\
WaterSplat-SLAM & \underline{0.432} & \underline{0.882} & \underline{0.350} & \underline{0.335} & \textbf{0.139} & 0.636 \\
Ours        & \textbf{0.422} & 0.958 & \textbf{0.157} & \textbf{0.182} & \underline{0.211} & \textbf{0.478} \\
\bottomrule
\end{tabular}
}
\end{table}

\subsection{Comparasion}
\subsubsection{Reconstruction Results Analysis}
Tab.~\ref{tab:reconstruction_results} reports the quantitative reconstruction results across six real-world underwater benchmarks. Among these, AquaFlow achieves state-of-the-art reconstruction performance on five datasets, outperforming existing off-line reconstruction and 3DGS-based SLAM methods by consistently improving PSNR and SSIM while reducing LPIPS. Although AquaFlow yields slightly lower metrics than UW-GS on SeaThru-NeRF, UW-GS is an off-line method that relies on substantially longer optimization time (Tab.~\ref{tab:time_efficiency}). Noticeably, SeaThru-NeRF is a small-scale benchmark where each scene contains only around 20 images focused on localized coral environments. This performance discrepancy further indicates that current off-line underwater methods tend to overfit to such small, single public benchmark datasets, thereby confirming the necessity and effectiveness of our extensive generalization evaluation across larger and more diverse real-world environments.

Fig.~\ref{fig:rendering} illustrates the qualitative reconstruction comparisons. As shown in the top three rows, owing to our explicit physical modeling of underwater light transport phenomena and the structured distance-conditioned Gaussian representation, AquaFlow achieves superior modeling of long-range objects as well as fine seafloor details and textures. The bottom three rows present reconstruction results in scenes with complex structural topologies (e.g., underwater caves and shipwreck ruins). Benefiting from our accurate appearance representation, the incorporated underwater geometric priors effectively constrain and enhance the structural geometry, enabling AquaFlow to faithfully preserve sharp boundaries and intricate structures where baseline approaches fail.

\subsubsection{Tracking Results Analysis}

Tab.~\ref{tab:tracking_results} presents quantitative trajectory localization performance. AquaFlow achieves the lowest overall mean error of 0.401\,m, consistently outperforming optical flow-based SLAM (DROID-SLAM), 3D vision foundation models (e.g., MASt3r-SLAM, VGGT-SLAM2), and underwater descriptor matching methods (UFEN-SLAM). While conventional baselines struggle with severe underwater visual degradation, our method leverages 3D underwater geometric priors to ensure robust pose estimation. Notably, AquaFlow achieves top tracking performance on four out of six real-world datasets, highlighting the strong generalization capability of our underwater 3D geometric prior across diverse underwater environments.

\subsubsection{Efficiency Analysis}

Tab.~\ref{tab:time_efficiency} evaluates the computational efficiency and memory footprint across six underwater benchmarks. We compare AquaFlow against the offline optimization baseline UW-GS as well as state-of-the-art streaming methods (ARTDECO and WaterSplat-SLAM). While delivering superior overall reconstruction quality compared to UW-GS, AquaFlow operates significantly faster, achieving substantial speedups in per-frame optimization time. Notably, this speed gap is even wider in practice, as our report excludes the heavy preprocessing runtime required by UW-GS for COLMAP pose estimation and Depth Anything V2\cite{yang2024depth} depth extraction. Compared to streaming baselines, AquaFlow maintains comparable or superior runtime efficiency while achieving significantly higher reconstruction accuracy. These efficiency gains stem from our lightweight LGPD network architecture and a medium-aware Gaussian initialization strategy, which precisely adds Gaussians where needed and properly initializes their attributes to avoid redundant computation.

\begin{table}[!t]
\centering
\scriptsize
\caption{\textbf{Ablation study on the FLSea-Canyons dataset.} We report PSNR, SSIM, and LPIPS. Higher PSNR/SSIM and lower LPIPS indicate better reconstruction quality.}
\label{tab:ablation}

\resizebox{0.85\linewidth}{!}{
\begin{tabular}{lcccc}
\toprule
Method & PSNR$\uparrow$ & SSIM$\uparrow$ & LPIPS$\downarrow$ & GS(k) \\
\midrule
Full model & \textbf{34.47} & \textbf{0.953} & \textbf{0.209} & 601.6\\
w/o UW finetuning & 33.41 & 0.943 & 0.247 & 608.3\\
w/o medium init. & 33.91 & 0.949 & 0.223 & \textbf{406.8}\\
w/o dist. embedding & 32.56 & 0.943 & 0.232 & 629.6\\
w/o LGPD & 33.92 & 0.951 & 0.214 & 653.2\\
\bottomrule
\end{tabular}
}

\end{table}

\begin{figure}[t]
\centering
\includegraphics[width=\columnwidth]{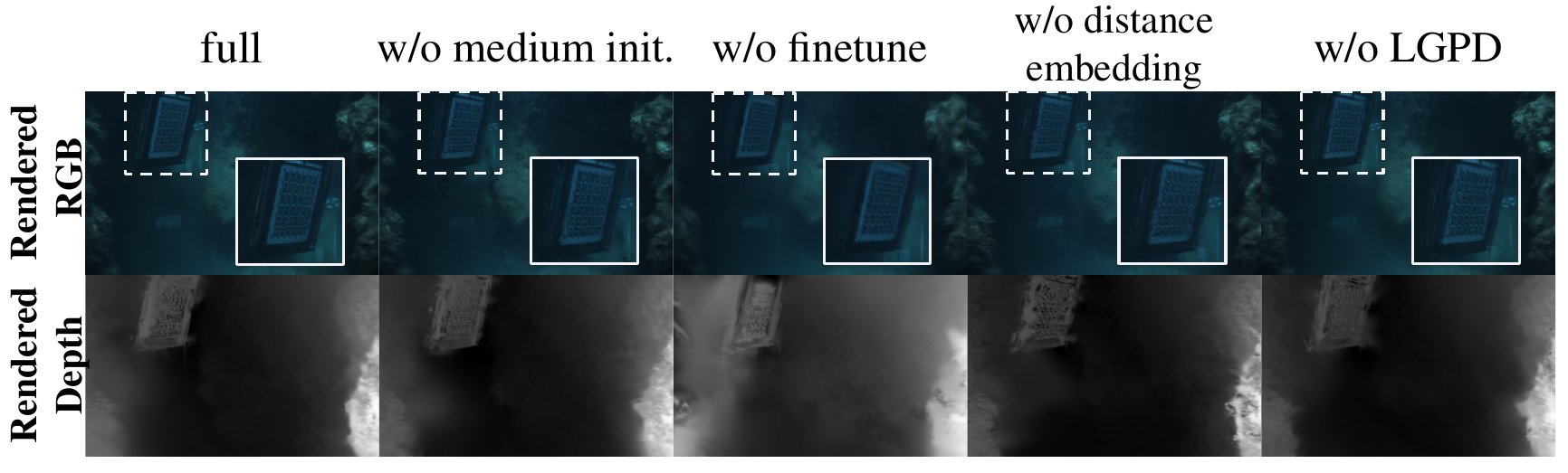}
\caption{\textbf{Qualitative ablation comparison on rendered color and depth maps.} 
Compared to baseline variants lacking specific modules (e.g., medium initialization, underwater fine-tuning, distance encoding, or LGPD), disabling any component introduces structural geometric errors or scene blurring. 
In contrast, our full model faithfully recovers both accurate 3D geometry and realistic underwater appearance.}
\label{fig:ablation_vis}
\end{figure}

\subsubsection{Ablation Study}

To evaluate the contribution of each key design, we conduct ablation studies on the FLSea-Canyons dataset (Tab.~\ref{tab:ablation} and Fig.~\ref{fig:ablation_vis}). Quantitatively, our full model achieves optimal performance across all metrics, whereas removing any core component leads to consistent performance drops. Specifically, we construct variant baselines by removing medium-aware initialization (\textit{w/o medium init.}), observation distance embeddings (\textit{w/o dist. embedding}), domain-adapted weights (\textit{w/o UW finetuning}), and replacing the LGPD module with an equivalent MLP (\textit{w/o LGPD}).

Qualitatively, Fig.~\ref{fig:ablation_vis} and Tab.~\ref{tab:ablation} visually confirm the individual contribution of each component in AquaFlow. Specifically, disabling medium-aware initialization (\textit{w/o medium init.}) yields sparse, under-dense initial seeds due to severe underwater attenuation and low contrast, ultimately leading to an insufficient modeling capability of Gaussian primitives and noticeable scene structural loss. Removing domain adaptation (\textit{w/o UW finetuning}) severely impairs the learned 3D geometric priors, resulting in corrupted depth maps with noisy boundaries and pronounced geometry collapse, which further directly degrades the final rendering quality. Meanwhile, omitting observation distance embeddings (\textit{w/o dist. embedding}) prevents the model from perceiving distance-dependent light scattering, causing apparent color shifts and contrast degradation in synthesized views. Finally, replacing the LGPD module with a standard MLP (\textit{w/o LGPD}) weakens the physical decoupling between local water turbidity and global illumination, leading to fine-detail blurring and loss of high-frequency textures. In contrast, our full model seamlessly integrates all modules to achieve the most visually pleasing and geometrically accurate reconstruction.

\section{Conclusion}

In this work, we present AquaFlow, a monocular 3D Gaussian Splatting streaming reconstruction framework tailored for complex underwater environments. By adapting a 3D vision foundation model to large-scale underwater data, AquaFlow achieves robust camera pose tracking and geometric prior estimation. For streaming mapping, we incorporate a medium-guided incremental Gaussian initialization strategy along with a streaming-compatible hybrid scene representation, effectively modeling range-dependent attenuation, scattering, and complex physical image formation phenomena. Extensive evaluations across 62 real-world underwater trajectories demonstrate that AquaFlow achieves state-of-the-art localization and reconstruction performance, outperforming strong offline and streaming baselines while maintaining high time efficiency. 

Building upon these capabilities, our future work will proceed in two directions. First, we plan to further enhance the zero-shot generalization of underwater visual foundation models by scaling up training datasets and further incorporating physics-inspired underwater priors into the training process. Second, we aim to extend our reconstruction framework to accommodate dynamic underwater environments, such as moving marine organisms, drifting floating particles, and flow-induced structural deformations (e.g., swaying vegetation or flexible cables), thereby broadening its applicability to a wider range of unconstrained underwater video inputs.

% \section*{Acknowledgments}
% This should be a simple subsubsection before the References to thank those individuals and institutions who have supported your work on this article.
{

\bibliographystyle{IEEEtran}
\bibliography{references}

@article{song2022optical,
  title={Optical imaging and image restoration techniques for deep ocean mapping: a comprehensive survey},
  author={Song, Yifan and Nakath, David and She, Mengkun and K{\"o}ser, Kevin},
  journal={Journal of Photogrammetry, Remote Sensing and Geoinformation Science (PFG)},
  volume={90},
  pages={243--267},
  year={2022},
  publisher={Springer}
}

@article{wolfl2019seafloor,
  title={Seafloor mapping--the challenge of a truly global ocean bathymetry},
  author={W{\"o}lfl, Anne-Cathrin and Snaith, Helen and Amirebrahimi, Sam and Devey, Colin W and Dorschel, Boris and Ferrini, Vicki and Huvenne, Veerle AI and Jakobsson, Martin and Jencks, Jennifer and Johnston, Gordon and others},
  journal={Frontiers in Marine Science},
  volume={6},
  pages={434383},
  year={2019},
  publisher={Frontiers}
}

@article{hu2023overview,
  title={Overview of underwater 3D reconstruction technology based on optical images},
  author={Hu, Kai and Wang, Tianyan and Shen, Chaowen and Weng, Chenghang and Zhou, Fenghua and Xia, Min and Weng, Liguo},
  journal={Journal of Marine Science and Engineering},
  volume={11},
  number={5},
  pages={949},
  year={2023},
  publisher={MDPI}
}

@article{qin2022survey,
  title={A survey on visual navigation and positioning for autonomous UUVs},
  author={Qin, Jiangying and Li, Ming and Li, Deren and Zhong, Jiageng and Yang, Ke},
  journal={Remote Sensing},
  volume={14},
  number={15},
  pages={3794},
  year={2022},
  publisher={MDPI}
}

@article{ferrera2019real,
  title={Real-time monocular visual odometry for turbid and dynamic underwater environments},
  author={Ferrera, Maxime and Moras, Julien and Trouv{\'e}-Peloux, Pauline and Creuze, Vincent},
  journal={Sensors},
  volume={19},
  number={3},
  pages={687},
  year={2019},
  publisher={MDPI}
}

@article{chemisky2021portable,
  title={A portable opto-acoustic survey solution for mapping of underwater targets},
  author={Chemisky, B and Nocerino, E and Menna, F and Nawaf, MM and Drap, P},
  journal={The International Archives of the Photogrammetry, Remote Sensing and Spatial Information Sciences},
  volume={43},
  pages={651--658},
  year={2021},
  publisher={Copernicus Publications G{\"o}ttingen, Germany}
}

@inproceedings{joshi2022high,
  title={High definition, inexpensive, underwater mapping},
  author={Joshi, Bharat and Xanthidis, Marios and Rahman, Sharmin and Rekleitis, Ioannis},
  booktitle={2022 International Conference on Robotics and Automation (ICRA)},
  pages={1113--1121},
  year={2022},
  organization={IEEE}
}

@inproceedings{wang2023real,
  title={Real-time dense 3d mapping of underwater environments},
  author={Wang, Weihan and Joshi, Bharat and Burgdorfer, Nathaniel and Batsosc, Konstantinos and Lid, Alberto Quattrini and Mordohaia, Philippos and Rekleitisb, Ioannis},
  booktitle={2023 IEEE International Conference on Robotics and Automation (ICRA)},
  pages={5184--5191},
  year={2023},
  organization={IEEE}
}

@article{bodenmann2017methods,
  title={Methods for quantitative studies of seafloor hydrothermal systems using 3D visual reconstructions},
  author={Bodenmann, Adrian and Thornton, Blair and Nakajima, Ryota and Ura, Tamaki},
  journal={Robomech Journal},
  volume={4},
  number={1},
  pages={22},
  year={2017},
  publisher={Springer}
}

@article{hernandez2016autonomous,
  title={Autonomous underwater navigation and optical mapping in unknown natural environments},
  author={Hern{\'a}ndez, Juan David and Isteni{\v{c}}, Klemen and Gracias, Nuno and Palomeras, Narcis and Campos, Ricard and Vidal, Eduard and Garcia, Rafael and Carreras, Marc},
  journal={Sensors},
  volume={16},
  number={8},
  pages={1174},
  year={2016},
  publisher={MDPI}
}

@article{kerbl20233d,
  title={3d gaussian splatting for real-time radiance field rendering.},
  author={Kerbl, Bernhard and Kopanas, Georgios and Leimk{\"u}hler, Thomas and Drettakis, George and others},
  journal={ACM Trans. Graph.},
  volume={42},
  number={4},
  pages={139--1},
  year={2023}
}

@article{li2024watersplatting,
  title = {{WaterSplatting}: Fast Underwater {3D} Scene Reconstruction Using Gaussian Splatting},
  author = {Li, Huapeng and Song, Wenxuan and Xu, Tianao and Elsig, Alexandre and Kulhanek, Jonas},
  journal = {arXiv preprint arXiv:2408.08206},
  year = {2024},
  url = {https://arxiv.org/abs/2408.08206}
}

@inproceedings{yang2024seasplat,
  title = {{SeaSplat}: Representing Underwater Scenes with {3D} Gaussian Splatting and a Physically Grounded Image Formation Model},
  author = {Yang, Daniel and Leonard, John J. and Girdhar, Yogesh},
  booktitle = {IEEE International Conference on Robotics and Automation},
  year = {2025},
  url = {https://arxiv.org/abs/2409.17345}
}

@inproceedings{jiang2025rusplatting,
  title = {{RUSplatting}: Robust {3D} Gaussian Splatting for Sparse-View Underwater Scene Reconstruction},
  author = {Jiang, Zhuodong and Wang, Haoran and Huang, Guoxi and Seymour, Brett and Anantrasirichai, Nantheera},
  booktitle = {British Machine Vision Conference},
  year = {2025},
  url = {https://arxiv.org/abs/2505.15737}
}

@article{xing2025uw3dgs,
  title = {{UW-3DGS}: Underwater {3D} Reconstruction with Physics-Aware Gaussian Splatting},
  author = {Xing, Wenpeng and Chen, Jie and Yang, Zaifeng and Lin, Changting and Dong, Jianfeng and Chen, Chaochao and Zhou, Xun and Han, Meng},
  journal = {arXiv preprint arXiv:2508.06169},
  year = {2025},
  url = {https://arxiv.org/abs/2508.06169}
}

@article{meuleman2025fly,
  title={On-the-fly reconstruction for large-scale novel view synthesis from unposed images},
  author={Meuleman, Andreas and Shah, Ishaan and Lanvin, Alexandre and Kerbl, Bernhard and Drettakis, George},
  journal={ACM Transactions on Graphics (TOG)},
  volume={44},
  number={4},
  pages={1--14},
  year={2025},
  publisher={ACM New York, NY, USA}
}

@article{zhang2025hi,
  title={Hi-slam2: Geometry-aware gaussian slam for fast monocular scene reconstruction},
  author={Zhang, Wei and Cheng, Qing and Skuddis, David and Zeller, Niclas and Cremers, Daniel and Haala, Norbert},
  journal={IEEE Transactions on Robotics},
  volume={41},
  pages={6478--6493},
  year={2025},
  publisher={IEEE}
}

@inproceedings{cheng2025outdoor,
  title={Outdoor monocular slam with global scale-consistent 3d gaussian pointmaps},
  author={Cheng, Chong and Yu, Sicheng and Wang, Zijian and Zhou, Yifan and Wang, Hao},
  booktitle={Proceedings of the IEEE/CVF International Conference on Computer Vision},
  pages={26035--26044},
  year={2025}
}

@inproceedings{murai2025mast3rslam,
  title     = {{MASt3R-SLAM}: Real-Time Dense SLAM with 3D Reconstruction Priors},
  author    = {Murai, Riku and Dexheimer, Eric and Davison, Andrew J.},
  booktitle = {Proceedings of the IEEE/CVF Conference on Computer Vision and Pattern Recognition},
  year      = {2025}
}

@article{maggio2026vggt,
  title={Vggt-slam: Dense rgb slam optimized on the sl (4) manifold},
  author={Maggio, Dominic and Lim, Hyungtae and Carlone, Luca},
  journal={Advances in Neural Information Processing Systems},
  volume={38},
  pages={129839--129867},
  year={2026}
}

@article{maggio2026vggt2,
  title={VGGT-SLAM 2.0: Real time Dense Feed-forward Scene Reconstruction},
  author={Maggio, Dominic and Carlone, Luca},
  journal={arXiv preprint arXiv:2601.19887},
  year={2026}
}

@inproceedings{levy2023seathru,
  title = {{SeaThru-NeRF}: Neural Radiance Fields in Scattering Media},
  author = {Levy, Deborah and Peleg, Amit and Pearl, Naama and Rosenbaum, Dan and Akkaynak, Derya and Korman, Simon and Treibitz, Tali},
  booktitle = {Proceedings of the IEEE/CVF Conference on Computer Vision and Pattern Recognition},
  year = {2023},
  url = {https://arxiv.org/abs/2304.07743}
}

@article{chen2025underwater,
  title={An underwater visual SLAM system with adaptive image enhancement},
  author={Chen, Gang and Du, Guoqiang and Yang, Chenguang and Xu, Yidong and Wu, Chuanyu and Hu, Huosheng and Dong, Fei and Zeng, Jinfeng},
  journal={Ocean Engineering},
  volume={326},
  pages={120896},
  year={2025},
  publisher={Elsevier}
}

@article{zheng2023real,
  title={Real-time GAN-based image enhancement for robust underwater monocular SLAM},
  author={Zheng, Ziqiang and Xin, Zhichao and Yu, Zhibin and Yeung, Sai-Kit},
  journal={Frontiers in Marine Science},
  volume={10},
  pages={1161399},
  year={2023},
  publisher={Frontiers Media SA}
}

@article{yang2023knowledge,
  title={Knowledge distillation for feature extraction in underwater VSLAM},
  author={Yang, Jinghe and Gong, Mingming and Nair, Girish and Lee, Jung Hoon and Monty, Jason and Pu, Ye},
  journal={arXiv preprint arXiv:2303.17981},
  year={2023}
}

@article{wang2025nerf,
  title={NeRF-Enhanced Visual--Inertial SLAM for Low-Light Underwater Sensing},
  author={Wang, Zhe and Zhang, Qinyue and Hu, Yuqi and Zheng, Bing},
  journal={Journal of Marine Science and Engineering},
  volume={14},
  number={1},
  pages={46},
  year={2025},
  publisher={MDPI}
}

@article{wang2025eum,
  title={Eum-slam: An enhancing underwater monocular visual slam with deep learning-based optical flow estimation},
  author={Wang, Xiaotian and Fan, Xinnan and Liu, Yueyue and Xin, Yuanxue and Shi, Pengfei},
  journal={IEEE Transactions on Instrumentation and Measurement},
  year={2025},
  publisher={IEEE}
}

@article{wu2025raem,
  title={RAEM-SLAM: A Robust Adaptive End-to-End Monocular SLAM Framework for AUVs in Underwater Environments},
  author={Wu, Yekai and Li, Yongjie and Luo, Wenda and Ding, Xin},
  journal={Drones},
  volume={9},
  number={8},
  pages={579},
  year={2025},
  publisher={MDPI}
}

@inproceedings{wang2025uw,
  title={UW-GS: Distractor-aware 3d gaussian splatting for enhanced underwater scene reconstruction},
  author={Wang, Haoran and Anantrasirichai, Nantheera and Zhang, Fan and Bull, David},
  booktitle={2025 IEEE/CVF Winter Conference on Applications of Computer Vision (WACV)},
  pages={3280--3289},
  year={2025},
  organization={IEEE}
}

@article{wang2026watersplat,
  title={WaterSplat-SLAM: Photorealistic Monocular SLAM in Underwater Environment},
  author={Wang, Kangxu and Zou, Shaofeng and Jiang, Chenxing and Dai, Yixiang and Chen, Siang and Shen, Shaojie and Wang, Guijin},
  journal={IEEE Robotics and Automation Letters},
  year={2026},
  publisher={IEEE}
}

@inproceedings{matsuki2024gaussian,
  title={Gaussian splatting slam},
  author={Matsuki, Hidenobu and Murai, Riku and Kelly, Paul HJ and Davison, Andrew J},
  booktitle={Proceedings of the IEEE/CVF conference on computer vision and pattern recognition},
  pages={18039--18048},
  year={2024}
}

@inproceedings{li2026artdeco,
  title={Artdeco: Toward high-fidelity on-the-fly reconstruction with hierarchical gaussian structure and feed-forward guidance},
  author={Li, Guanghao and Ren, Kerui and Xu, Linning and Zheng, Zhewen and Jiang, Changjian and Gao, Xin and Dai, Bo and Pu, Jian and Yu, Mulin and Pang, Jiangmiao},
  booktitle={The Fourteenth International Conference on Learning Representations},
  year={2026}
}

@article{deng2024compact,
  title={Compact 3d gaussian splatting for dense visual slam},
  author={Deng, Tianchen and Chen, Yaohui and Zhang, Leyan and Yang, Jianfei and Yuan, Shenghai and Liu, Jiuming and Wang, Danwei and Wang, Hesheng and Chen, Weidong},
  journal={arXiv preprint arXiv:2403.11247},
  year={2024}
}

@inproceedings{sandstrom2025splat,
  title={Splat-slam: Globally optimized rgb-only slam with 3d gaussians},
  author={Sandstr{\"o}m, Erik and Zhang, Ganlin and Tateno, Keisuke and Oechsle, Michael and Niemeyer, Michael and Zhang, Youmin and Patel, Manthan and Van Gool, Luc and Oswald, Martin and Tombari, Federico},
  booktitle={Proceedings of the Computer Vision and Pattern Recognition Conference},
  pages={1680--1691},
  year={2025}
}

@misc{wildflow_sweet_corals_indo_2025,
    author       = { Sergei Nozdrenkov and Mujiyanto and Raymon Rahmanov Zedta and Fakhrurrozi and Triona Barker and Muhammad Dziki Rahman and Andrias Steward Samusamu and Fathur Rochman and Leigh Peters and Dr Rita Rachmawati and Dr Ofri Johan and Dr Jamie Craggs and Prof Michael Sweet },
    title        = { sweet-corals (Revision bd316dc) },
    year         = 2025,
    url          = { https://huggingface.co/datasets/wildflow/sweet-corals },
    doi          = { 10.57967/hf/5162 },
    publisher    = { Hugging Face }
}

@inproceedings{schoenberger2016sfm,
    author={Sch\"{o}nberger, Johannes Lutz and Frahm, Jan-Michael},
    title={Structure-from-Motion Revisited},
    booktitle={Conference on Computer Vision and Pattern Recognition (CVPR)},
    year={2016},
}

@article{randall2023flsea,
  title   = {{FLSea}: Underwater Visual-Inertial and Stereo-Vision Forward-Looking Datasets},
  author  = {Randall, Yelena and Treibitz, Tali},
  journal = {arXiv preprint arXiv:2302.12772},
  year    = {2023},
  url     = {https://arxiv.org/abs/2302.12772}
}

@inproceedings{wang2020tartanair,
  title={Tartanair: A dataset to push the limits of visual slam},
  author={Wang, Wenshan and Zhu, Delong and Wang, Xiangwei and Hu, Yaoyu and Qiu, Yuheng and Wang, Chen and Hu, Yafei and Kapoor, Ashish and Scherer, Sebastian},
  booktitle={2020 IEEE/RSJ International Conference on Intelligent Robots and Systems (IROS)},
  pages={4909--4916},
  year={2020},
  organization={IEEE}
}

@inproceedings{alvarez2023mimir,
  title={MIMIR-UW: A multipurpose synthetic dataset for underwater navigation and inspection},
  author={{\'A}lvarez-Tu{\~n}{\'o}n, Olaya and Kanner, Hemanth and Marnet, Luiza Ribeiro and Pham, Huy Xuan and le Fevre Sejersen, Jonas and Brodskiy, Yury and Kayacan, Erdal},
  booktitle={2023 IEEE/RSJ International Conference on Intelligent Robots and Systems (IROS)},
  pages={6141--6148},
  year={2023},
  organization={IEEE}
}

@inproceedings{leroy2024grounding,
  title={Grounding image matching in 3d with mast3r},
  author={Leroy, Vincent and Cabon, Yohann and Revaud, J{\'e}r{\^o}me},
  booktitle={European conference on computer vision},
  pages={71--91},
  year={2024},
  organization={Springer}
}

@inproceedings{akkaynak2018revised,
  title     = {A Revised Underwater Image Formation Model},
  author    = {Akkaynak, Derya and Treibitz, Tali},
  booktitle = {Proceedings of the IEEE/CVF Conference on Computer Vision and Pattern Recognition},
  pages     = {6723--6732},
  year      = {2018}
}

@inproceedings{joshi2019experimental,
  title={Experimental comparison of open source visual-inertial-based state estimation algorithms in the underwater domain},
  author={Joshi, Bharat and Rahman, Sharmin and Kalaitzakis, Michail and Cain, Brennan and Johnson, James and Xanthidis, Marios and Karapetyan, Nare and Hernandez, Alan and Li, Alberto Quattrini and Vitzilaios, Nikolaos and others},
  booktitle={2019 IEEE/RSJ International Conference on Intelligent Robots and Systems (IROS)},
  pages={7227--7233},
  year={2019},
  organization={IEEE}
}

@article{wang2004image,
  title={Image quality assessment: from error visibility to structural similarity},
  author={Wang, Zhou and Bovik, Alan C and Sheikh, Hamid R and Simoncelli, Eero P},
  journal={IEEE transactions on image processing},
  volume={13},
  number={4},
  pages={600--612},
  year={2004},
  publisher={IEEE}
}

@inproceedings{zhang2018unreasonable,
  title={The unreasonable effectiveness of deep features as a perceptual metric},
  author={Zhang, Richard and Isola, Phillip and Efros, Alexei A and Shechtman, Eli and Wang, Oliver},
  booktitle={Proceedings of the IEEE conference on computer vision and pattern recognition},
  pages={586--595},
  year={2018}
}

@inproceedings{teed2021droidslam,
  title     = {{DROID-SLAM}: Deep Visual SLAM for Monocular, Stereo, and RGB-D Cameras},
  author    = {Teed, Zachary and Deng, Jia},
  booktitle = {Advances in Neural Information Processing Systems},
  volume    = {34},
  pages     = {16558--16569},
  year      = {2021}
}

@inproceedings{murai2024mast3r,
  title = {{MASt3R-SLAM}: Real-Time Dense {SLAM} with {3D} Reconstruction Priors},
  author = {Murai, Riku and Dexheimer, Eric and Davison, Andrew J.},
  booktitle = {Proceedings of the IEEE/CVF Conference on Computer Vision and Pattern Recognition},
  year = {2025},
  url = {https://arxiv.org/abs/2412.12392}
}

@article{yang2024depth,
  title={Depth anything v2},
  author={Yang, Lihe and Kang, Bingyi and Huang, Zilong and Zhao, Zhen and Xu, Xiaogang and Feng, Jiashi and Zhao, Hengshuang},
  journal={Advances in Neural Information Processing Systems},
  volume={37},
  pages={21875--21911},
  year={2024}
}

}

\vfill

\end{document}